\documentclass[preprint, 5p, authoryear, twocolumn, times]{elsarticle}

\journal{Neural Networks}

\biboptions{authoryear}

\usepackage[cmex10]{amsmath}

\usepackage[utf8]{inputenc} 
\usepackage{graphicx}
\usepackage{supertabular} %
\usepackage{booktabs} 
\usepackage[safe]{tipa} 
\usepackage[T1]{fontenc} 
\usepackage{bm} 
\usepackage[ruled, linesnumbered]{algorithm2e} 
\usepackage{stfloats} 
\usepackage{graphicx}
\usepackage{threeparttable} 
\usepackage{multirow} 
\usepackage{caption} 
\usepackage{array} 
\usepackage{siunitx} 
\usepackage{amsmath,amssymb} 

\newcommand{\figref}[1]{Figure \ref{#1}}
\newcommand{\tabref}[1]{Table \ref{#1}}
\newcommand{\secref}[1]{Section \ref{#1}}
\newcommand{\equaref}[1]{Equation \ref{#1}}
\newcommand{\algref}[1]{Algorithm \ref{#1}}

\usepackage{color, soul} 

\begin{document}
\begin{frontmatter}

\title{A Neural Network Architecture for Learning Word-Referent Associations in Multiple Contexts}

\author{Hansenclever F. Bassani}
\author{Aluizio F. R. Araujo}
\address{Centro de Informática - Universidade Federal de Pernambuco, Recife-PE, Brazil, 50.740-560}

\begin{abstract}
This article proposes a biologically inspired neurocomputational architecture which learns associations between words and referents in different contexts, considering evidence collected from the literature of Psycholinguistics and Neurolinguistics. The multi-layered architecture takes as input raw images of objects (referents) and streams of word's phonemes (labels), builds an adequate representation, recognizes the current context, and associates label with referents incrementally, by employing a Self-Organizing Map which creates new association nodes (prototypes) as required, adjusts the existing prototypes to better represent the input stimuli and removes prototypes that become obsolete/unused. The model takes into account the current context to retrieve the correct meaning of words with multiple meanings. Simulations show that the model can reach up to 78\% of word-referent association accuracy in ambiguous situations and approximates well the learning rates of humans as reported by three different authors in five Cross-Situational Word Learning experiments, also displaying similar learning patterns in the different learning conditions.
\end{abstract}


\end{frontmatter}



\section{Introduction}
\label{sec:introduction}

    Language is surely a vital and distinctive trait of human beings. Even though  language acquisition by young children is an active research topic in cognitive sciences, a number of open issues persist, despite the achievements of the field. For instance, we do not know exactly how humans acquire the meaning of words, an essential part of the language acquisition process. In this article, we propose a word learning model composed of a set of neural modules, or schemes \citep{Arbib2008}, that simultaneously compete and cooperate to perform higher-level tasks. The model was proposed considering the evidence brought by the literature of neurolinguistics and psycholinguistics about the characteristics of the word learning capabilities displayed by humans. With that, the proposed model is able to simulate multiple statistical characteristics displayed by humans when they learn new words.

	We assume that word learning may be studied disregarding the interference of other aspects of language acquisition, such as the acquisition of grammar, semantics, and pragmatics. Therefore, according to \cite{Bloom2002}, in order to learn the meaning of a word, an individual must learn three different elements: (i) the concept or meaning of the word (\textbf{referent}); (ii) the sound or lexical representation of the word (\textbf{label}); and (iii) the \textbf{association} between referent and label. Each of these challenging tasks will be addressed in this article.

	A classic example \citep{Quine1960} illustrates the difficulties that children and foreign language learners have to handle to correctly match words and referents. When a native speaker of an unknown language sees a white rabbit and pronounces ``gavagai'', one might understand this as clear evidence that the word ``gavagai'' means rabbit. However, such sound could also mean ``white'', ``furry'', ``food'', ``let's go hunting'' or even something completely unrelated with rabbit, such as ``it is going to rain''. The expression ``gavagai'' could even be a composition of two or three words with their own meanings.

	One possible strategy to address the problem described by \cite{Quine1960}, is known as ``cross-situational word learning'' (CWSL) \citep{Yu2007}. In this type of learning, the words would not be learned after a single exposure. The learning process would consider information from multiple learning trials. Thus, a learner who is unable to decide unambiguously the meaning of a word after a single trial would form a new knowledge subject to be further strengthened or weakened upon new evidence.

    Currently, we can argue that word learning requires a set of cognitive abilities that are not yet fully understood \citep{Bloom2002}, such as theory of mind (the ability to simulate and understand the thought of others), concept acquisition, and fast mapping (the ability to associate referents and labels with few, or even one trial). In this article, we focus on the last two abilities of this list.

    Concept acquisition may be seen as the ability to recognize and group similar referents together so that the category itself (concept) could be further associated with a label. \cite{Harnad2005} points out that ``To Cognize is to Categorize'' and \cite{Perlovsky2006} describes the mind as a hierarchy of multiple layers of concept-models, from simple elements like edges or moving dots to more abstract concept-models of objects, relationships, complete scenes, and so on.

    The proposed model is compatible with these views because it defines the learning tasks mentioned above as a subspace clustering problem \citep{Kriegel2005,Bassani2015,Hu2018}, in which the cluster prototypes capture the concept-models. At the current state, the model focuses on the lower levels of the concept-model hierarchy mentioned by Perlovsky, learning the referents, labels, and their associations for concrete nouns that can be depicted in static images, such as \textit{chair}, \textit{table}, and \textit{pen}, in their different usage contexts (basic concept-models). The model learns such elements incrementally by creating new prototype nodes as required, adjusting the existing prototypes to better represent the auditory and visual input stimuli or removing prototypes that become obsolete/unused.

    To achieve this, we specify a neurocomputational architecture composed of four layers: (i) the first layer extracts the perceptions from raw visual data (the referents) and auditory data (the labels); (ii) the second layer creates a more suitable representation for labels and referents; (iii) the third layer recognizes the current context and; (iv) the fourth layer creates the associations between labels and referents in the different contexts in which they are used, thus forming the prototypes representing the basic concept-models learned by the model.

    In order to evaluate the proposed model, we simulate the CSWL experiments carried out with human beings by \cite{Yu2007}, \cite{Yurovsky2013}, and \cite{Trueswell2013}. These experiments provide sound evidence on the operation of word learning mechanisms. Any model aiming to represent the functioning of these learning mechanisms must be able to reproduce to some extent the world learning patterns described in the following paragraphs.

	\cite{Yu2007} designed experiments to evaluate the abilities of humans in acquiring correct word-referent pairings and they have found compelling evidence that adult humans are able to learn label-referent pairings through CSWL. In their experiments, the stimuli consisted of slides containing 2, 3, or 4 pictures of unusual objects paired with 2, 3, or 4 pseudowords presented in the auditory form. These artificial words were generated by a computer program using standard phonemes in English. In this case, the label-referent pairs were formed by single and unique objects randomly chosen, used in three different training conditions of ambiguity.

\begin{figure}
\begin{minipage}{\linewidth}
	\fbox{\includegraphics[width=0.95\linewidth]{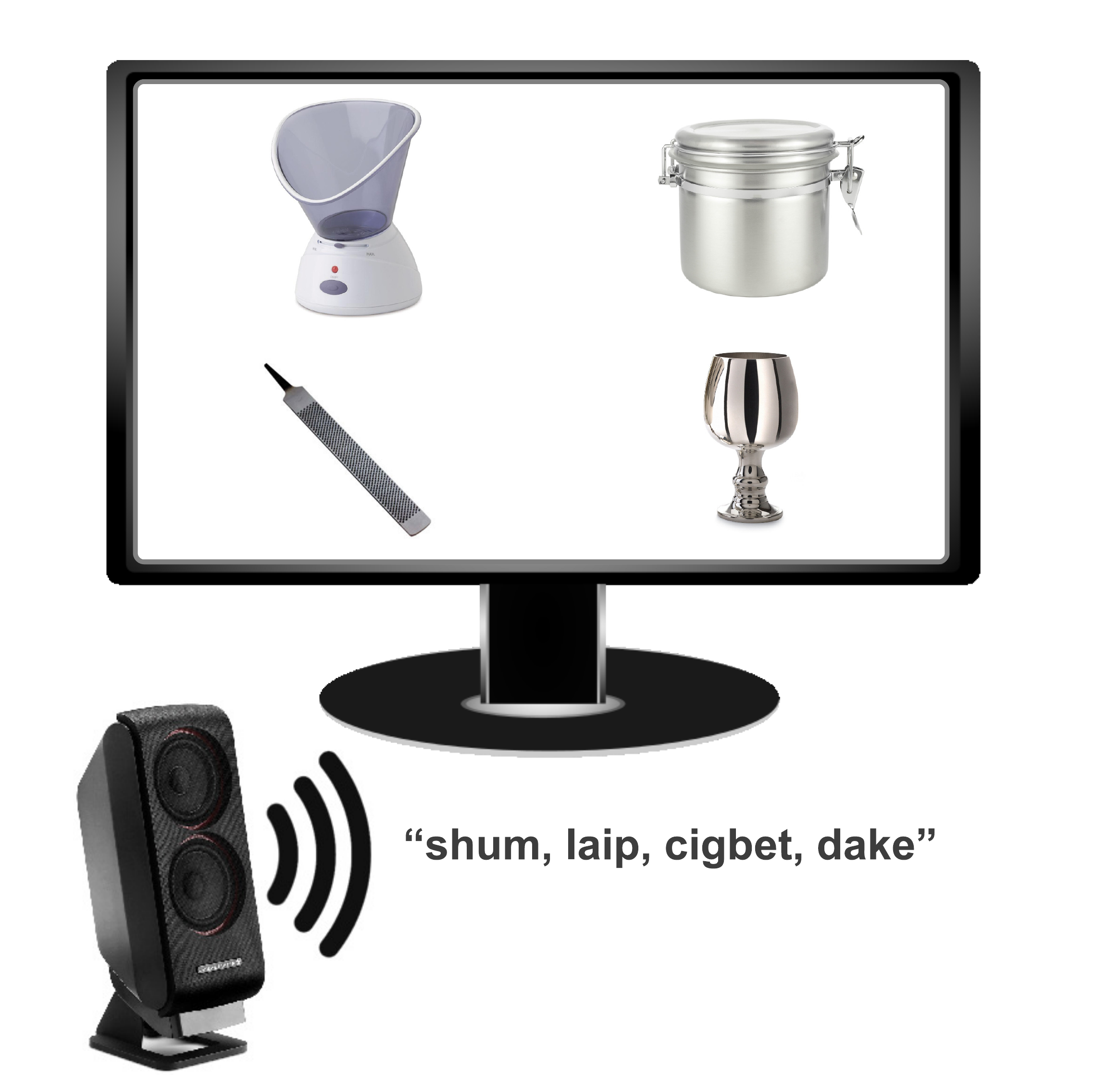}}
	\captionof{figure}{Illustration of a trial in the 4x4 condition. The pictures of four objects (referents) are shown in the monitor while the sound of four pseudowords is presented auditorily over the speakers (labels).}
	\label{fig:crosssituacional}
\end{minipage}
\end{figure}

	The training conditions differ only in the number of labels and referents simultaneously presented to the subjects. \figref{fig:crosssituacional} illustrates a 4x4 condition, in which four objects (referents) were presented simultaneously on the screen, while the sound of 4 pseudowords (labels) were heard from the speakers. The results showed that the individuals were able to discover on average more than 16 out of the 18 pairs in the 2x2 condition and more than 13 out of the 18 pairs in the 3x3 condition. 

    \cite{Yurovsky2013} expanded the previous experiment including situations in which labels could be associated with more than one referent. They were interested in evaluating if there was competition occurring in the learning process and if it was local (among referents presented in the same trial) or global (among referents presented in different trials). Their results suggested that global competition is most likely to occur.
    
    The computational models proposed in the literature for CSWL can be divided into two categories \citep{Yu2007}: the Hypothesis-Testing Models, in which the learner maintains a list of hypothesized pairings to be further confirmed or rejected due to a mutual exclusivity constraint and the Associative Models, a basic form of Hebbian learning which strengths associations between observed word-referent pairs.

    \cite{Trueswell2013} designed experiments to compared the two hypotheses and their results suggested that subjects did not keep track of multiple candidate meanings for each label, hence, according to the authors, such experiments weaken the hypothesis that humans employ some kind of statistical learning of the word-referent pairings.

    Current studies have focused on comparing these two modeling approaches in terms of how well they fit experimental data, but no consensus has emerged yet. For instance, \cite{Kachergis2017} found that an associative model which includes competition between familiarity and uncertainty biases reproduces better the individual and combined effects of frequency and contextual diversity on human learning. \cite{Khoe2019} found that this associative model better captures the full range of individual differences and conditions when learning is cross-situational, although the hypothesis testing approach outperforms it when there is no referential ambiguity during training.
    
    The model proposed in this article differs from these studies by focusing in dealing with real-world data (raw images and phoneme sequences) and in employing a neural network architecture that can be used to simulate models of both categories, though in the present work the associative approach was considered.
    
    The obtained results show that the proposed model is able to replicate the patterns of CSWL presented by humans. Additionally, the proposed model was also tested in scenarios in which there was ambiguity about the correct word-referent parings, with more than one association. We show that the model can take into account the context to solve ambiguity and choose the correct referent for ambiguous words.

    The following sections of this article are structured as follows: \secref{sec:evidences} discusses the Associationism theory and presents the experimental evidence on word-referent associations. \secref{sec:previous-models} describes correlated models for language acquisition. \secref{sec:proposed-model} presents the proposed modular architecture for replicating the CSWL experiments while \secref{sec:som} and \secref{sec:ART2WithContext} detail the two neural network models employed in the learning tasks, LARFDSSOM, and ART2 with Context. \secref{sec:simulations} describes the CSWL experiments performed by \cite{Yu2007}, \cite{Yurovsky2013}, and \cite{Trueswell2013} along with the simulations carried out with the proposed model for replicating them. Finally, \secref{sec:conclusaoexperimentos} discusses and summarizes the main conclusions drawn from the obtained results.

\section{Associationism and Experimental Evidence About How Humans Learn Word-Referent Associations}
\label{sec:evidences}

    Associationism is one of the most widely held theories of learning, appearing since John Locke (1700). According to it, learning is based on sensibility to covariation of the human brain. \cite{Richards1986} proposed that children could learn the meaning of a word by repeatedly associating its verbal label with their perceptual experience at the time that the label is used. For those perceptual properties that repeatedly co-occur with the label, the association strengthens.

    We can find several pieces of evidence supporting Associationism in word learning. For instance, children's first words often refer to things that they can see and touch; words are learned best in conditions in which an associative match would be easier to make. Additionally, the results of cross-situational word learning show that adults can learn word-referent associations with repeated co-occurrence. However, Associationism cannot explain all the observed word learning phenomena. Below, we list the most significant points collected by \cite{Bloom2002} against a pure associationist theory of word learning.

\begin{enumerate}
\item Associationism requires that label and referent are simultaneously present in the environment. However, studies show that about 30-50\% of the time a word is used, young children are not attending to the object the adult is talking about \citep{Collins1977, Harris1983, Bunce2017}.
\item Associationism predicts that before children have enough data to retrieve the right associations they would often make mapping errors unless they wait until having collected strong statistical evidence. However, it was observed that in certain situations, children can learn a new word even after a single exposition \citep{Markson1997,Frank2014}.
\item Association between labels and perceptions does not explain how children learn labels of more abstract referents that they cannot see or touch. A significant number of children's words refer to abstract conceptual categories such as ``morning'' or ``day'' \citep{Nelson1993,Feijoo2017}.
\end{enumerate}

The view of the authors of this work is that the capability of statistical association is necessary, though not sufficient, for word learning, and it can serve as a basis for other higher cognitive functions. We are interested in verifying how well we can model the human word learning behavior in cross-situational word learning with a modular neural network that learns statistical correlations.

This modular network was built considering evidence collected from the literature of Psycholinguistics, Neurolinguistics and organized them in a modular architecture which presents similarities to those employed in Computational Linguistics \citep{Allen1994}. 

Below, we present the evidence that we collected from the literature, separated by their field. In section \secref{sec:proposed-model} we present the proposed architecture and discuss how each piece of evidence was taken into account in its specification.

\subsection{Evidence from Psycholinguistics}
\label{sec:psychoevidence}

\noindent
\textbf{Cross-situational word learning}: There is plenty of work \citep{Yu2007,Yurovsky2013, Trueswell2013, Bunce2017} showing that human adults can robustly figure out the correct word-referent associations in ambiguous learning situations, in which the correct mapping of a word to an intended referent cannot be guaranteed. The learning rates and patterns presented by humans in different conditions of ambiguity provide valuable information for evaluating word learning models.

\noindent
\textbf{Correcting feedback is not a requirement}: Correcting feedback may help learning, however, children do not require it to learn word meanings. \cite{Lieven1994}, reviews works showing that there are cultures in which adults do not even speak directly to children until they are using words in a meaningful manner. This suggests a computational model considering unsupervised or reinforcement learning.

\noindent
\textbf{``New word, new object'' preference}: Studies suggest that children are biased to consider that each word is associated with a single referent \citep{Kagan1981,Markman1988}. Therefore, if they are presented with a new word they will prefer to associate it with a currently unlabeled referent. This is also known as ``mutual exclusivity''.

\noindent
\textbf{Object categorization can be biased by labels}: Most labels are associated not with a singular object but with a category of similar objects (that share certain properties). For instance, the word ``car'' refers to a set of different types of vehicles that share certain features. \cite{Plunkett2008} show that the choice of what labels are presented for children as naming new objects can affect how they categorize these objects, biasing them to create certain categories that they would not create otherwise. \cite{Mayor2010} created a neurocomputational model that successfully reproduced this behavior for simulated data.

\noindent
\textbf{Different features are relevant for each category}: The properties young children attend to when categorizing a novel entity depend on its type (object versus a non-solid substance) \citep{Soja1991}, plant or rock \citep{Keil1994}, real or toy monkey \citep{Carey1995}, animal or tool \citep{Becker1991}. This suggests the employment of subspace clustering methods in the categorization of items to form the referent concepts. In subspace clustering, certain attributes can be more relevant than others for each category, and an item may belong to more than one category. For instance, consider the categorization of a red hexagon. This object belongs to different categories depending on the features that are taken into account. Regarding its color, it belongs to the category of red objects, while regarding its shape it belongs to the category of hexagonal objects. Finally, it belongs to a third category when taking both features into account.

\noindent
\textbf{Fast Mapping}: Other studies \citep{Carey1978,Dollaghan1985,Heibeck1987,Rice1990,Markson1997} show that children and adults can learn word-referent associations after a few exposures (even one), without any explicit training or feedback, and even without any explicit act of naming.

\noindent
\textbf{Context can affect retrieved memories}: Brainerd and Reyna (1998, 2008) have shown that in experiments in which a list of words with a shared central meaning are presented for subjects to memorize, after the memorization, the subjects are induced to recognize as having seen on the list words related with this central meaning even when they were not on the list (false memories). These experiments suggest that the contextual meaning formed during the pattern presentations plays an important role for memorization and is taken into account during recognition \citep{Matzen2009}. This behavior was modeled and reproduced by \cite{Araujo2010} with a modular neural network.

\subsection{Evidence from Neurolinguistics}
\label{sec:neuroevidence}

\noindent
\textbf{Hierarchical perceptual processing}: Sensory information is processed to extract information that is relevant to the individual (perceptions), through innate or self-adaptive processes, probably in inferior cortical regions such as the visual cortex \citep{Miikkulainen2005} and auditory cortex \citep{Pasley2012}. Superior cortical areas, such as V5 and the posterior parietal cortex integrate information to form more complete perceptions \citep{Udesen1992,Born2005}.

\noindent
\textbf{Mirror neurons}: Certain neurons respond to correlated perceptual information from different modalities, such as verbal, visual and motor information about the same action or event, as observed in the sensory-motor cortex \citep{Rizzolatti2004,Pulvermuller2005}.

\noindent
\textbf{Context recognition}: Hippocampus and amygdala keep a historical record of the input stimuli, forming a kind of context \citep{Fletcher1997,Aggleton1999}.

\noindent
\textbf{Topographic-preserving input mapping}: Nearby neurons in the brain respond to inputs with similar features as in certain areas of the brain where topographic maps are formed, especially in the primary motor, visual, and somatosensory cortical areas \citep{Haykin1998,Spitzer1999,Miikkulainen2005}.

\section{Previous Language Acquisition Models Based on Self-Organizing Maps}
\label{sec:previous-models}

Considering that children are able to acquire language without explicit feedback, several language acquisition models are based on unsupervised learning methods. Self-Organizing Maps \citep{Kohonen1982} and Adaptive Resonant Theory (ART) \citep{Grossberg1976a,Grossberg1976b} are two of the most prominent unsupervised learning neural networks. ART was employed for modeling human behavior in the task of memorization of word lists \citep{Pacheco2004, Araujo2010}, while several computational models for word learning are based on SOM \citep{Ritter1989, Miikkulainen1997, Plunkett1992, Plunkett1997, Li2004, Silberman2007, Li2007}. Refer to \cite{Li2013} for a review of SOM-based language acquisition models.

\cite{Ritter1989} applied SOM to capture the semantic structure of words. Their pioneer work showed that implicit categories in the linguistic environment can be recognized by SOM.

\cite{Guenther1996} have shown that a SOM fed with formant representation of different phonemic categories can simulate the perceptual magnet effect \citep{Kuhl1991}, an effect characterized by a warping of the perceptual space near central phonemic category, that allows certain sound categories to be considered as more similar to each other than to those patterns further away from the center.

The associative hypothesis is explicitly modeled by Hebbian learning in DISLEX, DevLex, and DevLex II models. The basic idea is that the activation of co-occurring lexical and semantic representations in each map leads to an adaptive formation of associative connections between them.

\cite{Miikkulainen1997} introduced the DISLEX model to simulate dyslexia and aphasia. The model was the first to connect different SOMs through associative links. Each SOM represents a different type of linguistic information, such as phonological, orthographic and semantic. DISLEX has also been shown to be able to simulate patterns of bilingual language recovery in aphasic patients \citep{Kiran2013}.

Following this structure, two models, DevLex \citep{Li2004} and DevLex II \citep{Li2007}, were proposed to simulate children's early lexical development. Instead of employing maps with a fixed structure, in the DevLex family, new nodes are inserted in the map when required, to improve the accuracy of learning. DevLex has been shown to model patterns of lexical confusion as a function of word density and semantic similarity, simulating age-of-acquisition effects while learning a growing lexicon. DevLex II has been shown to simulate several empirical phenomena, including patterns of vocabulary spurt, the relationship between comprehension and production, fast mapping, lexical category development and, lexical overextension.

\cite{Silberman2007} employed a single layer SOM for simulating the associations between words and concepts in a semantic network that extracts semantic information from the CHILDES database \citep{Macwhinney2010}. The model was able to replicate learning patterns such as the effects of semantic priming that indicates faster response when recognizing a word semantically related to the information in the episodic memory, than when recognizing unrelated words.

\cite{Mayor2010} presented a model for simulating fast mapping in early word learning. Their model included two SOMs, one fed with visual input representing artificial objects and the other fed with acoustic input representing words. The connections between the two SOMs were also adjusted by Hebbian learning. The model displayed learning patterns of early lexical category development, such as the tendency to attribute to a new object a known name of another object in the same category.

Despite the acknowledgeable achievements of these models, none of them was designed to replicate the CSWL experiments, which is an excellent source of data about word-referent associations. In this regard, \cite{Yu2012} described and compared two competing types of models for CSWL: Hypothesis-Testing Models and Associative Models. In Associative Models \citep{Yu2007}, the representation is a large word-object matrix in which each cell contains the associative strength between one word and one object and a basic form of Hebbian learning is employed to strength associations between observed word-referent pairs. In the Hypothesis-Testing Models \cite{Medina2011,Trueswell2013}, the learner maintains a list of hypothesized pairings (a single hypothesis for each word) to be further confirmed or rejected due to a mutual exclusivity constraint. Both types of models were shown to be able to replicate the patterns of CSWL and the main conclusion of the authors was that it is necessary to look at the components of models to understand how they contribute to overall learning.

 Such models, however, were not modular and were not developed to work with real-world input data, such as images and sounds. This limits their ability to replicate the details of experiments carried out with humans. The next section describes the modular architecture we proposed to address those issues.

\section{Proposed Modular Architecture}
\label{sec:proposed-model}

\begin{figure}
\begin{minipage}{\linewidth}
    \fbox{\includegraphics[width=0.97\linewidth]{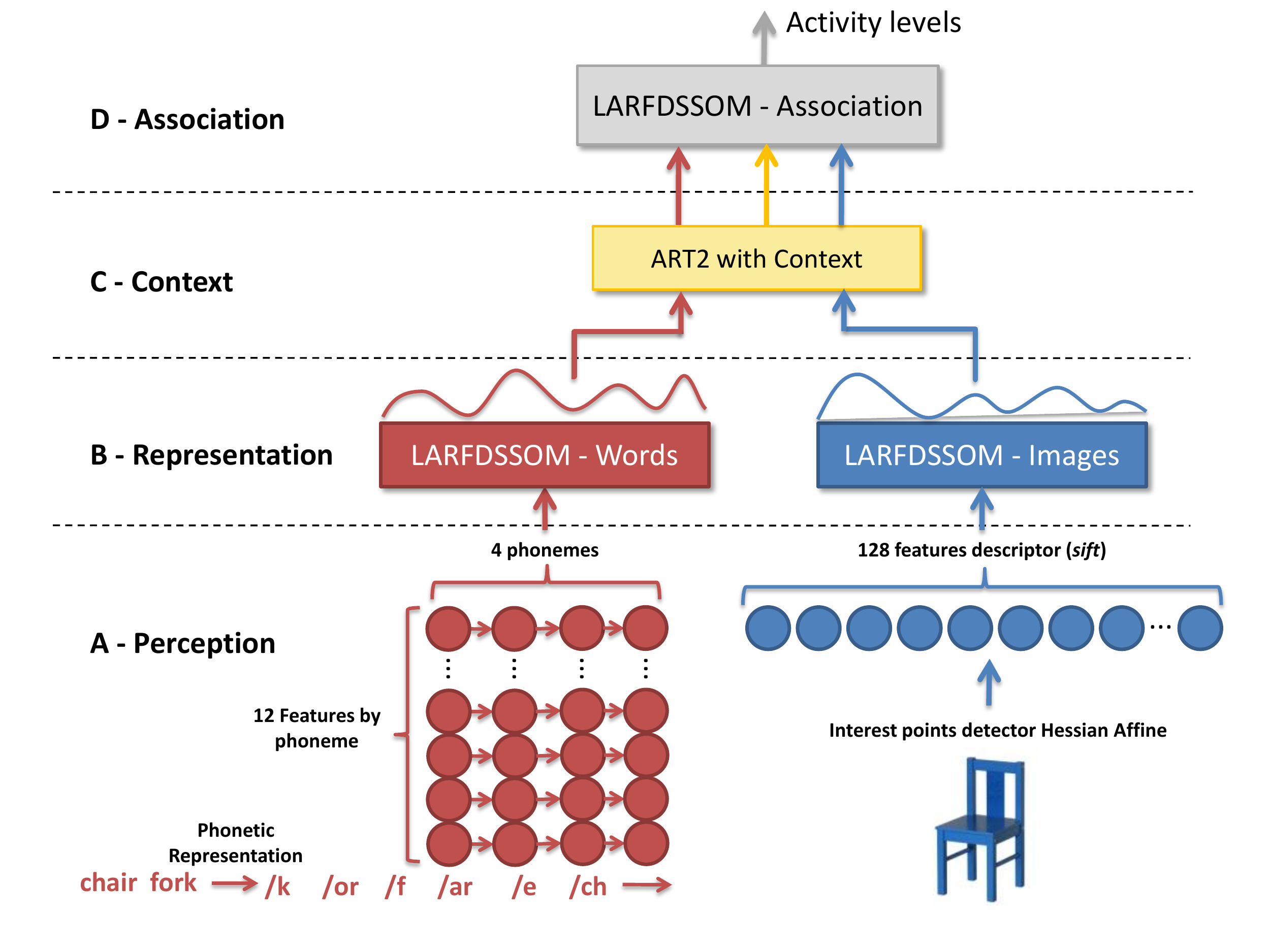}}
    \captionof{figure}{Illustration of the processing layers of the model. A - Perception Acquisition; B - Representation; C - Context formation and recognition; and D - Association and context-dependent recognition.}
    \label{fig:modulosexperimentos}
\end{minipage}
\end{figure}

\figref{fig:modulosexperimentos} illustrates the proposed architecture, which is stratified in four layers. The first two layers are comprised of parallel modules that are specialized for each kind of stimuli (auditory or visual), while the third and fourth layers present one module each performing multisensory integration. Below we present a general description of each layer:

\begin{enumerate}[A]
\item -- Perception: It extracts relevant information (perceptions) from the sensory data. The sensory-perception mapping modules present in this layer are specialized for each kind of input. The auditory module extracts phonemes from a sound (or from a text, for convenience), while the visual module extracts descriptions of interest points from image patches.

\item -- Representation: It consolidates perceptions that are distributed in space or time, creating a representation that is suitable for understanding a given stimulus. This layer contains representation modules specialized for each type of stimulus (visual or auditory). For instance, an isolated phoneme may carry little meaning, however, a sequence of phonemes could represent a word or a lemme (temporal consolidation). Similarly, in the visual processing, the description of a small patch of an image may carry little meaning, however, the description of a set of patches can carry information enough to represent an object or a scene (spacial consolidation).

\item -- Context: This layer contains the context module that receives the multisensory perceptions as input, accumulates sequences of these inputs, and clusters them to form a "temporal context" that can be recognized afterward. The context recognition is important, for instance, to disambiguate the meaning of homophone/homograph words, such as \textit{mouse} (animal or computer device). The recognized context is forwarded to the next layer together with the inputs received.

\item -- Association: The module in this layer, associates (or integrates) the perceptions of words, visual objects and contexts. This association is achieved by the means of perception clustering. Therefore, each cluster represents an association. For instance, each word can be associated with different meanings that occur in different contexts by being represented in more than one cluster. In the same way, a visual object can be associated with more than one word by being represented in more than one cluster. For instance, the object \textbf{car} can be associated with the words \textit{car} and \textit{vehicle}, in two different clusters. 
\end{enumerate}

\figref{fig:modulosexperimentos} indicates the learning models employed in each module, as well as how the information flows through the whole architecture. In the following subsections, we describe each module in more detail. The learning models are described afterward.

\subsection{Sensory-Perceptive Mapping Modules}
\label{sec:sensoryperceptivemapping}

	In the CSWL experiments, visual and auditory stimuli are simultaneously presented to the subjects, as depicted in \figref{fig:crosssituacional}. In the proposed model, these two kinds of stimulus are processed in parallel in the first layer to produce a numeric representation of the perceptions as output as described in the following subsections.

\subsubsection{The Auditory Sensory-Perceptive Mapping}
\label{sec:auditoryrepresentation}

    The auditory input data consists of a stream of text representing the name of each object displayed on the scream. For instance, the string: "mixer, canister, rasp, goblet", would describe the objects in \figref{fig:crosssituacional}. 

    In order to obtain a numeric representation of the auditory data, we followed a procedure similar to that described by \cite{Araujo2010}. First, we convert each word to its respective phonetic representation. This step employs the CMU Pronouncing Dictionary (CMUdict) \citep{Lenzo2007}. Therefore, the example above is translated into: "K AE N AH S T ER, R AE S P, G AA B L AH T, M IH K S ER", in which, each phoneme is represented by its ARPAbet symbol, separated by spaces.
    
    Afterward, each phoneme is translated into a vector of 12 real values ranging from -1 to +1 (see \tabref{tab:PhoneticSymbolsRepresentation} in \ref{sec:apnumericrepphon}). This numeric representation was built considering the place of pronunciation of each phoneme in the International Phonetic Alphabet (IPA) charts for vowels and consonants, encoding specific features for vowels (4 of them) and for consonants (8 of them). Therefore, when a vowel is represented, the features for consonants are set to zero, and when a consonant is represented, the features of vowels are set to zero. The rationale behind this procedure is to obtain similar representations for phonemes with similar sounds.

	Finally, the representation of any sequence of words is a list of vectors, each vector describing the characteristics of one phoneme in the sequence. This list represents the perception output by the Auditory Sensory-Perceptive Mapping.
    
\subsubsection{The Visual Sensory-Perceptive Mapping}
\label{sec:visualrepresentation}

	The extraction of visual perceptions consists of detecting and describing numerically the parts of the object present in the image. In this article, we follow the literature of Unsupervised Object Discovery \citep{Weber2000,Tuytelaars2010,Kinnunen2012}, and we use the Scale Invariant Feature Transform (SIFT) to detect Points of Interest (POIs) and describe each POI as a vector of 128 values \citep{Lowe1999}, called ``POI descriptor''. These POI descriptors are normalized by an L2 normalization.

    In this module, each object in the screen is represented by a list of descriptors of the POIs detected and described by SIFT. For instance, in the 4x4 condition exemplified in \figref{fig:crosssituacional}, we have four objects on the screen that will result in four lists of POI descriptor vectors, one list per object. These lists represent the perception output by the visual Sensory-Perceptive Mapping.

	The outputs of both Sensory-Perceptive Mapping modules in Layer I are, then, fed as inputs to the respective Representation Modules in Layer II.

\subsection{Representation Modules}
\label{sec:representation modules}

	A feature vector produced by both modules described above, considered in isolation, is not enough to identify the auditory or visual elements. For instance, one phoneme is not enough to identify a word, analogously, the descriptor of one POI of an image cannot identify an object. Therefore, it is necessary to compose the information from several feature vectors to properly describe an element of interest, thus allowing its recognition.
    
    The basic idea employed in this module is to build a Bag-of-Features (BoF) representation, by determining and stringing the features distributed in space or time. This approach was used for Unsupervised Visual Object Discovery (UVOC) from Images by \cite{Tuytelaars2010} and \cite{Kinnunen2012}. It derives from the Bag-of-Words (BoW) approach, a way to represent text \citep{Salton1986} for categorization tasks. The BoF approach consists of two steps: first, similar features are clustered to create a dictionary of features called ``codebook'', in which, the number of clusters determines the size of the features vector produced to represent the objects. After creating this dictionary, the objects are described by counting the number of features mapped in each cluster, thus, forming a histogram of occurrence, which is usually normalized.

    In \cite{Tuytelaars2010}, several clustering methods and types of histogram normalization were evaluated. The authors concluded that when there is one object category per image, even k-means yields good results, being outperformed only by spectral clustering. \cite{Kinnunen2012} considered SOM to be a viable alternative of clustering method for BoF. The authors obtained similar results to those presented by \cite{Tuytelaars2010}. However, they found SOM to be more robust to the type of normalization applied to the histogram.

    Instead of the traditional SOM, we employ LARFDSSOM in the representation layer to generate the codebook. LARFDSSOM is a suitable method for this task because it is capable of subspace clustering and it employs a locally weighted distance metric to adjust the relevances of the input dimensions. This is an important property when the input data present high dimensionality, since it is able to identify, for instance, which kinds of image patches are relevant for determining each object category and its associated phonetic representation. A detailed description of LARFDSSOM is provided in Section \ref{sec:som}.

    The representation module maps were pre-trained to learn a codebook, forming 28 clusters in the phonetic representation map and 37 clusters in the visual representation map. This training has occurred in advance since these maps represent the previous knowledge that each individual has about the phonetic structure of its native language and about the basic perceptual elements necessary to recognize objects.

\subsection{Context Module}
\label{sec:context}

This module should associate a context with each newly received input, in a way to distinguish the same stimulus presented under distinct contexts, and also, to approximate different inputs when presented in similar contexts. In the brain, this role is played by the hippocampus where, several recurrent connections are observed in the cortical regions of memorization, hence recurrent neural networks seem to be a suitable approach. Hence, we applied the ART2 with context described in Section \ref{sec:ART2WithContext}.

The visual and auditory representations are given as input to the ART2 With Context, which recognize the current context or create a new context if necessary. The outputs of the context module consist of the visual and auditory inputs, unchanged, associated with the context representation recognized by ART2 with Context, and stored by its context units, $UC$.

\subsection{Association Module}
\label{sec:association}

The Association Module takes as input the three outputs of the context module, visual, auditory and contextual information to associate them. In this article, this task is also carried out by a LARFDSSOM. The map computes the activation of all existing nodes and the node with the highest activation, the winner node, represents the best association found. If its activation is above the threshold parameter, $a_t$, this node is updated to slightly modify the previous association. Otherwise, a new node is inserted in the map to represent a new association learned as it is presented in its inputs.

    It is worth pointing out that, as the nodes on the map are updated, they learn which input dimensions are relevant. This allows the nodes to take into account only the aspects of the visual, auditory and contextual information that present a certain level of correlation. For instance, if a word occurs frequently with the same sound in several different contexts, the node can learn that the context is irrelevant for this association. In another example, if certain aspects of the image correlate with a certain sound while others do not, the uncorrelated aspects are taken as irrelevant. 

    In our simulations, the LARFDSSOM was initialized with a single neuron randomly positioned in the input space and no limit was applied to the number of nodes created so that the network could grow as much as required to represent the associations found. The output of the association module is the activation of the winner node. If this value is above the threshold parameter, $a_t$, it indicates that the pattern presented by the inputs of the network was recognized, thus, the visual, contextual, and auditory information are considered associated and, the higher this value, the stronger the association made by the map is. This allows us to compare object-sound associations in different contexts and to identify the strongest association.

    During the recognition phase of the cross-situational word-learning simulations, all pairings of objects and sounds are presented as input for the model and the pair with the highest activation is considered as the strongest association made from the network.

\subsection{How the Evidence was Taken Into Account} 

Each piece of evidence collected in the literature and described in Sections \ref{sec:psychoevidence} and \ref{sec:neuroevidence} was somehow taken into account in the proposition of the architecture, as indicated below: 

\noindent
\textbf{Cross-situational word learning}: The proposed model was designed to replicate the CSWL experiments, while keeping the main aspects of the structure of previous SOM-based language acquisition models.

\noindent
\textbf{Correcting feedback is not a requirement}: The proposed model was developed based on unsupervised learning models, therefore it does not require correcting feedback.

\noindent
\textbf{``New word, new object'' preference}: Though this was not evaluated in our experiments, the similarity based competition employed in the learning model used in the association layer (LARFDSSOM), makes that stimuli significantly different from what was previously seen (novel stimuli) tend to be stored on new associations nodes.

\noindent
\textbf{Object categorization can be biased by labels}: The proposed architecture was specially designed to take this into account by making both labels and referents as inputs to the association layer. This allows labels to affect the categorization of referents, by making their representations more similar/different. This is also true for the contextual information.

\noindent
\textbf{Different features are relevant for each category}: This is a feature of LARFDSSOM, which learns the relevance of each input dimension for each category during the self-organization process.

\noindent
\textbf{Fast Mapping}: LARFDSSOM can learn new associations in one shot.

\noindent
\textbf{Context can affect retrieved memories}: In the proposed architecture, the current context is recognized and affects the information stored and retrieved, since it is part of the representation sent for the association layer.

\noindent
\textbf{Hierarchical perceptual processing}: This inspired the layered architecture proposed, which takes raw sensory data as input, extracts perceptions, converts it to a more suitable representation which is fed to the context formation layer, and finally, forwarded to the association layer.

\noindent
\textbf{Mirror neurons}: The nodes in the association layer perform the multisensory integration and can be activated by information of different modalities, similarly as the mirror neurons.

\noindent
\textbf{Topographic-preserving input mapping}: This inspired the employment of a SOM-based model with a topographic-preserving characteristic in layers B and D.

The following two sections provide details about the implementation of LARFDSSOM (employed in representation and association layers) and ART2 with Context (employed in the Context Layer). All the source-code and datasets produced in Perception Layer are available online \footnote{\label{fn:github}Available on GitHub: https://github.com/hfbassani/word-referent-association}.

\section{Subspace Clustering with Self-Organizing Maps}
\label{sec:som}

The Self-Organizing Map (SOM) proposed by \cite{Kohonen1982}, is a neural network trained with unlabeled data (unsupervised learning). It maps a high-dimensional data into a lower dimensional (usually bi-dimensional) grid of $N$ nodes (or neurons), compressing information while preserving the topological relationships of the original data.

The following characteristics of SOM are worth highlighting here:

\begin{itemize}
\item It creates an abstraction and a simplified representation of the input data distribution \citep{Haykin1998}. Each node can be seen as a prototype representing similar input data.
\item Its topological properties correlate with what is observed in the sensory processing regions of the brain, where the input stimuli are represented in topologically ordered neural maps \citep{Miikkulainen2005}. In particular, sensory inputs such as tactile \citep{Kaas1983}, visual \citep{Hubel1962,Hubel1977}, and acoustic \citep{Suga1985} inputs are mapped to different areas of the cerebral cortex in a topologically orderly manner.
\item SOM-based models were applied to a variety of problems involving sensory processing, including voice recognition and image processing \citep{Kangas1991, Venkateswarlu2011, Abdelsamea2015, Chen2017};
\end{itemize}

  These characteristics have made SOM a good candidate for modeling the processing of perceptions. However, as we mentioned in the previous section, traditional clustering algorithms (SOM included) are not adequate to create abstract representations in the form of perceptions, because they weight equally all input dimensions and because they map each input stimuli to a single cluster. These limitations prevent SOM from being able to correctly cluster this kind of data and create prototypes that represent the several possible abstractions associated with the same stimulus, as in the example of the \textit{red hexagon} given above. Therefore, other SOM-based subspace clustering methods that address these limitations are considered here.

The Dimension Selective Self-Organizing Map (DSSOM) \citep{Bassani2012} was one step towards making SOM adequate for subspace clustering. By using a weighted Euclidean distance (\equaref{eq:distKangas}) to compare samples and prototypes it is able to adjust the relevance of each dimension to determine the winning node for each grid node. Thus, the model allows the weight of some dimensions to be even zeroed so that these dimensions do not influence the selection of data clustered by a given node. The adjustment of these weights is done adaptively during self-organization process.

\begin{equation}
	[D_\omega(\textbf{x}, \textbf{c}_j)]^2 = \sum_{i=1}^m{\omega_{ji}^2(x_i - c_{ji})^2}
	\label{eq:distKangas}
\end{equation}
where $\textbf{x}$ is an input stimulus, $\textbf{c}_j$ is the $j$-th prototype on the map, and $\omega_{ji}^2 \in [0,1]$ is the weighting factor that the $j$-th prototype applies to the $i$-th input dimension.

These weighting factors are estimated from the variance of the input patterns clustered by each node on the grid. The higher the variance, the lower its weighting factor is. Moreover, DSSOM allows more than one node to win for a given input stimulus, so that, nodes that apply a set of weighting factors different from those considered by the previous winners can also group that stimulus.

   DSSOM presented solid results, comparable to or better than previous subspace clustering methods from the data mining field. However, the fixed topology of DSSOM ($N\times{}N$ grid) requires strong knowledge about the data, and may not adequately represent the neighborhood topology of clusters that live in different subspaces. This issue was addressed in the map described in the next section, which is the method that we have chosen to employ in the proposed model for learning word-referent associations.

\subsection{Local Adaptive Receptive Field Dimension Selective Self-Organizing Map - LARFDSSOM}

   LARFDSSOM \citep{Bassani2015} preserves the main characteristics of SOM and DSSOM. However, in LARFDSSOM the nodes are not organized in a fixed grid. Instead, it introduces a time-varying structure with a mechanism that inserts new nodes into the map whenever the winner node is not similar enough to the current input pattern. In order to achieve this, it defines an activation function (\equaref{eq:activation}), inversely related to the distance presented in \equaref{eq:distKangas} and a threshold parameter ($a_t$). When the activation of the winner node in response to an input pattern is below this threshold, a new node is inserted into the map, at the position of the input pattern.

\begin{equation}
		ac(D_\omega(\textbf{x}, \textbf{c}_j), \boldsymbol\omega_j) = \frac{1}{1 + D_\omega(\textbf{x}, \textbf{c}_j)/(\|\boldsymbol\omega_j\|^2 + \epsilon)}
\label{eq:activation}
\end{equation}
where $\epsilon$, is a small value to avoid division by zero, $\|\boldsymbol{\omega_j}\|$ is the norm of the relevance vector, and $D_\omega(\textbf{x}, \textbf{c}_j)$ is the weighted distance function shown in \equaref{eq:distKangas}.

	The relevance vector is computed as an inverse function of the average distance of each node to the input patterns that it clusters, $\bm\delta_s$, i.e., the greater is the average distance in a dimension, the smaller is the respective relevance (\equaref{eq:larfdssom_update}).

\vspace{-0.3cm}
\begin{equation}
\begin{aligned}
	\textbf{c}_j(n+1) &= \textbf{c}_j(n) + e(\textbf{x} - \textbf{c}_j(n)),\\
	\bm\delta_j(n+1) &= (1-e\beta) \bm\delta_j(n) +  e\beta(|\textbf{x} - \textbf{c}_j(n)|),\\
    \omega_{ji} &= \begin{cases}
  \displaystyle \frac{1}{1+\exp{\left(\frac{\delta_{jimean}-\delta_{ji}}{sl(\delta_{jimax} - \delta_{jimin})}\right)}} & \text{if } \delta_{jimin} \neq \delta_{jimax}\\
   1       & \text{otherwise}
 \end{cases}
\end{aligned}
\label{eq:larfdssom_update}
\vspace{0.2cm}
\end{equation}
where $e$ is the learning rate given by: $e = e_b$ if $j$ is the winner node and $e = e_n$ if $j$ is a neighbor of the winner node, $\delta_{jimax}$, $\delta_{jimin}$ and $\delta_{jimean}$ are respectively, the maximum, the minimum, and the mean of the components of the distance vector $\bm\delta_j$ and $e_b$, $e_n$, $sl$, $\beta$ $\in$ $[0,1]$ are parameters.

\begin{algorithm}[ht]
\small
\caption{Self-Organization Phase}
\label{alg:organizacaoLARFDSSOM}
	Initialize parameters $a_t$, $lp$, $nwins$, $maxcomp$... \;
	Initialize the map with one node with $\textbf{c}_j$ initialized at the first input stimulus, $\bm\delta_j \leftarrow \bm{0}$, $\bm\omega_s \leftarrow \bm{1} $ and $wins_j \leftarrow 0$\;
	Initialize the variable $nwins \leftarrow 1$\;
	\ForEach{input stimulus ($\emph{\textbf{x}}$)}{
		Present $\textbf{x}$ to the map\;
		Compute the activation of all nodes (\equaref{eq:activation})\;
		Find the winner $s$ with the highest activation ($a_s$)\;
		\uIf {$a_s < a_t$ and $N < N_{max}$} {
			Create new node $j$ setting: $\bm{c}_j \leftarrow \textbf{x}$, $\bm\delta_j \leftarrow \bm{0}$, $\bm\omega_j \leftarrow \bm{1}$ and $wins_j \leftarrow lp \times nwins$\;
			Setup the neighborhood of node $j$\;
		}
		\Else{
			Update the vectors $\textbf{c}$, $\bm\delta$, and $\bm\omega$  of the winner and of its neighbors (\equaref{eq:larfdssom_update})\;
			Set $wins_s \leftarrow wins_s + 1$\;		
		}
		\If {$nwins = maxcomp$} {
			Remove nodes with $wins_j < lp \times maxcomp$\;
			Update the connections of the remaining nodes\;
			Reset the number of wins of the remaining nodes: $wins_j \leftarrow 0$\;
			$nwins \leftarrow 0$\;
		}
		$nwins \leftarrow nwins + 1$\;
	}
\end{algorithm}
   
    Also, in LARFDSSSOM, nodes that do not cluster a minimum percentage ($lp$) of the input patterns are periodically removed from the map (every $maxcomp$ competitions). Additionally, the neighborhood connects only nodes that take into account a similar subset of the input dimensions. 

    The operation of the map comprises three phases: organization, convergence and clustering phase. In the organization phase, the nodes compete to cluster each new input pattern, so that the winner and its neighbors are updated to approximate it and new nodes are created whenever the most activated node does not reach the threshold $a_t$. The convergence phase is similar to the organization phase, with the exception that node insertion is not allowed. Finally, in the clustering phase, the consolidated map is not changed anymore, being used only for clustering.

    In this article, for simulating the learning process of a subject going through the CSWL experiments, we employ the organization phase (shown in Alg. 1) without limiting the number of nodes in the map and with nodes being updated as per \equaref{eq:larfdssom_update}, while the convergence phase is not used.
  
    The clustering phase (shown in Alg. 2) is used for testing what the simulated subjects have learned to recognize.

\begin{algorithm}[ht!]
\small
\caption{Clustering with LARFDSSOM}
\label{alg:clustering}
\ForEach{input pattern ($\emph{\textbf{x}}$) in the dataset}{
	Present $\textbf{x}$ to the map\;
	Compute the activation of all nodes (\equaref{eq:activation})\;
	Find the winner $s$ with the highest activation ($a_s$)\;
	\eIf {$a_s \geq a_t$}
	{
		\Repeat {$a_s < a_t$}
		{
			Assign $\textbf{x}$ to the cluster of the winner node $s$\;
			Find the next winner $s$ disregarding the previous winners\;
		}
	} {
    	$\textbf{x}$ was not recognized\;
    }
}
\end{algorithm}
\section{Context Formation and Recognition with ART2}
\label{sec:ART2WithContext}

Since words can have different meanings in different context, taking context into account when recognizing words is a fundamental task in word learning.  In this work, we employ for this task a neural network called ART2 with Context \cite{Araujo2010}, based on ART2 \cite{Carpenter1987} which is a model from the Adaptive Resonant Theory. Such an unsupervised incremental learning is capable of grouping patterns, associates stimuli of different natures, adjusts the degree of similarity of the grouped patterns, works with plasticity and stability, and presents some plausibility. \cite{Araujo2010} adapted ART2 by inserting context units with recurrent connections. These context units aim to store a history of the input patterns and make this context affect both pattern search and recognition phases.

\begin{figure}[hb!]
	\centering
	\includegraphics[width=\linewidth]{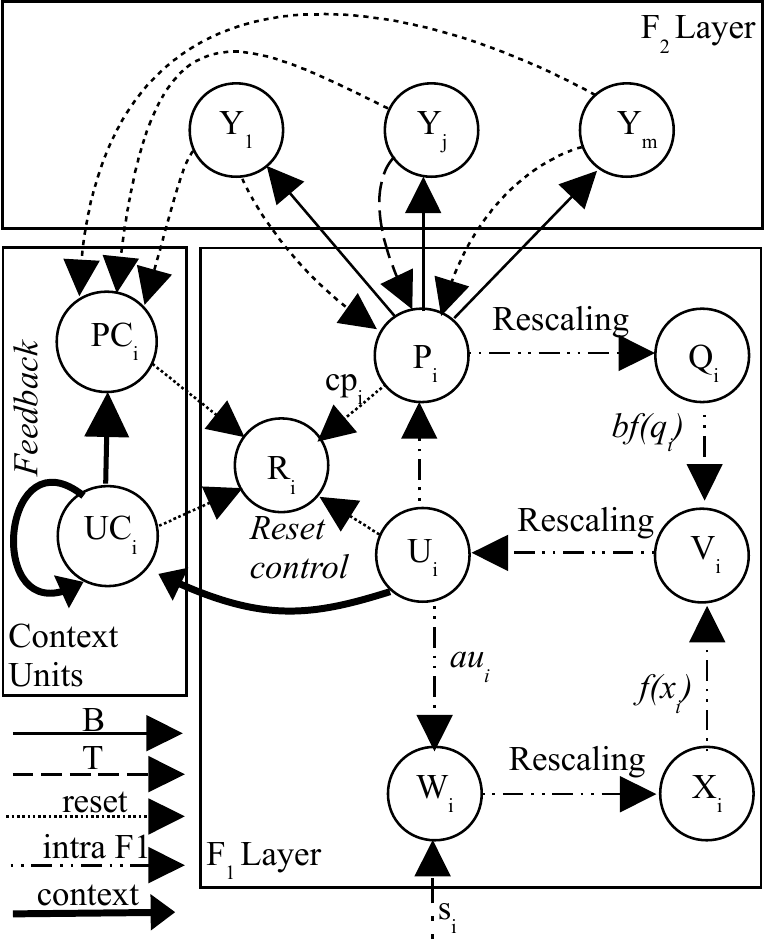}
	\caption{Architecture of ART2 with Context, composed of two layers: $F_1$ is the input layer, $F_2$ is the output layer; and the context units: $UC$, with a feedback loop, responsible for creating the context representation, and the $PC$ units serving as an interface between $F_1$ and $F_2$ layers.}
	\label{fig:ContextModule}
	\vspace{-10pt}
\end{figure}

The ART2 with Context, \figref{fig:ContextModule}, present the same input ($F_{1}$) and output ($F_{2}$) layers of ART2, however, context units $UC$ and $PC$ with recurrent connections were added to the model. Each $UC_{i}$ unit contains a kind of average of the input values. Each $U_{i}$ unit stores the intensity of the occurrence of a property in the input pattern, in the internal representation of ART2 network, i.e., properly rescaled and with noise suppression. Each $UC_{i}$ unit receives two connections: the new input pattern from $U_{i}$ and a feedback from itself with its own previously stored value. This feedback has a $back$ parameter which controls the weight of the previous value of each $UC_{i}$ unit. At the end of the presentation of a sequence of stimuli, it is expected that the context formed and stored in $UC$ units approximates an average representation of similar stimuli present in the sequence. The $PC$ units serve as an interface between $F_{1}$ and $F_{2}$ layers, and they have a role equivalent to the $P$ units of the original ART2 model.

\algref{alg:ART2Context} presents all the steps needed for training the ART2 with Context included in the proposed model, for which the parameters are:

$\eta$: number of nodes in the $F_{1}$ layer. It is equal to the number of semantic features.
$a$ and $b$: fixed weights in $F_{1}$. We set $a$ and $b$ = 10.
$c$: fixed weights used by the reset condition in [0,1] interval.
$d$: activation of the winner unit in $F_{2}$ within the [0,1] interval. The value 0.9 was used.
$e$: parameter to avoid division by zero when the norm of a vector is zero. The value 0.0001 was used.
$\theta$: parameter of noise suppression, typically $1/\sqrt{n}$. The input vector components with values lower than $\theta$ will have their values taken to zero.
$\alpha$: learning rate. Used value: 0.001.
$\rho$: surveillance parameter. To determine the number of groups to be formed. Values in [0.7,1] interval produce effective control over the number of groups formed. Used value: 1.
$epochs$: maximum number of epochs, we set it to 1.
$n_{iter}$: maximum number of iterations. Used value: 1
$back$: weight of the context in the interval [0,1]. Used value: 0.9.
$cw$: influence rate of the contextual information over the reset mechanism, inside [0,1] interval. Used value: 0.
$d_{ctx}$: effect equivalent to $d$ used for the context units. Used value: 0.9.
$\alpha_{ctx}$: context learning rate: Used value: 0.8.
The variables $p_{i},q_{i},r_{i},s_{i},u_{i},v_{i},w_{i},x_{i},y_{i},uc_{i},pc_{i},$ are the $i$-th elements of the vectors \textbf{P}, \textbf{Q}, \textbf{R}, \textbf{S}, \textbf{U}, \textbf{W}, \textbf{X}, \textbf{Y}, \textbf{UC}, and \textbf{PC}.
	J: the node in $F_{2}$ with higher activation.
	reset: indicates if the winner node in $F_{2}$ layer cannot learn the presented pattern.
	T: the top-down matrix of weights.
	B: the bottom-up matrix of weights.
	The $f(x)$ function is defined as:
	\[
		f(x) = 
		\begin{cases} 
		x & \text{if } x \geq \theta \\
		0 & \text{if } x < \theta
		\end{cases}
	\]

\begin{algorithm}[ht!]
	\footnotesize
	\caption{Training ART2 with Context.}
	\label{alg:ART2Context}
	 Initialize: $a$, $b$, $c$, $d$, $e$, $\alpha$, $\theta$, $\rho$, $epochs$, $n_{iter}$, $n$, $back$, $cw$, $d_{ctx}$, $\alpha_{ctx}$\;
	\For{$epochs$}{
		\For {$\text{each input stimulus } s_i$}{
			 {Initialize activations in $F_1$ layer:}\\
			 $u_i = 0\text{; } w_i = s_i\text{; } p_i = 0\text{; } q_i = 0$\text{; }		 $x_i = s_i / (e + ||s||)\text{; } v_i = f(x_i)$\;
			 {Update activations in $F_1$ Layer:}\\
			 $u_i = v_i / ( e + ||v|| )\text{; } w_i = s_i + au_i\text{; } p_i = ui$\;
			 $x_i = w_i / ( e + ||w|| )\text{; } q_i = p_i / ( e + ||p|| )\text{; } v_i = f(x_i) + bf(q_i)$\;
			 {Propagate values to $UC$:}\\
			 $uc_i = (back)(uc_i) + (1-back)f(u_i)$\;
			 {Rescale the context units:}\\
			 $uc_i = uc_i/(e+||uc||)$\;
			 {Propagate the context values to $PC$:}\\
			 $pc_i = uc_i$\;
			 {Update activations in $F_2$ layer:}\\
			 $y_j = (1-cw)\sum_{i} b_{i,j} p_i + cw\sum_{i} (b_{i+n,j} p_i)$\;
			 $reset = true$\;
			\While{reset}{
				{Find the unit in $F_2$ with highest activation $y_J$:}\\
				$y_J = max[y_j], 1\leq j\leq q$\;
				\If {$y_J = -1$} {
					 $J = \text{an unused unit}$\;
					 $reset = false$\;
				}
				\eIf{reset}{
					 $ui = vi / ( e + ||v|| )\text{; } p_i = u_i + dt_{J,i} \text{; } pc_{i+n} = t_{J,i+n}$\;
					 $r_i = (u_i+cp_i+cw)/(e + ||u|| + c||p|| + cw||pc||)$\;
					\eIf{$||r||<(\rho-e)$}{
						 $reset = true, y_J = -1$\;
					}{
						 $reset = false, w_i = s_i + au, x_i = w_i / (e+||w||)$\;
						 $q_i = p_i/(e+||p||), v_i = f(x_i) + b f(q_i)$\;
					}
				}
				{
					\For{$n_{iter}$}{
						 {Update the weights of the winner unit $J$:}\\
						 $t_J i = \alpha d u_i + [ 1 + \alpha d (d - 1)]t_{J,i}$\;
						 $b_i J = \alpha d u_i + [ 1 + \alpha d (d - 1)]b_{i,J}$\;
						 $t_{J,i+n} = \alpha_{ctx} d_{ctx} uc_i + [ 1 + \alpha_{ctx} d_{ctx} (d_{ctx} - 1)]t_{J,i+n}$\;
						 $b_{i+n,J} = \alpha_{ctx} d_{ctx} uc_i + [ 1 + \alpha_{ctx}d_{ctx} (d_{ctx} - 1)]b_{i+n,J}$\;
						 {Rescale the updated vectors:}\\
						 $t_{J,i} = t_{J,i} / ||t_{J}||$\text{; } $b_{i,J} = b_{i,J} / ||b_{J}||$\;
						 $t_{J,i+n} = t_{J,i+n} / ||t_{J}||$\text{; }$b_{i+n,J} = b_{i+n,J} / ||b_{J}||$\;
						 {Update activations in $F_1$ Layer:}\\
						 $u_i = v_i / ( e + ||v|| )$\text{; } $w_i = s_i + au_i$\text{; } $p_i = ui + d t_{Ji}$\;
						 $x_i = w_i / ( e + ||w|| )$\text{; } $q_i = p_i / ( e + ||p|| )$\;
                         $v_i = f(x_i) + bf(q_i)$\;
					}
				}
			}
		}
	}
\end{algorithm}

	The training algorithm (\algref{alg:ART2Context}) consists of the following: After the variable initializations (line 1) a loop is executed for each training epoch. For each input pattern the activations of the units in layers \textbf{U}, \textbf{W}, \textbf{P}, \textbf{Q}, \textbf{X}, and \textbf{V} are initialized (line 5) and updated to reflect the effects of the input pattern (lines 7 and 8). Then the values computed are propagated to the context units \textbf{UC} (line 10) and the new values are rescaled (line 12) and copied to \textbf{PC} units (line 14). Next, values stored on \textbf{P} and \textbf{PC} units are propagated to the $F_{2}$ layer, where a competition occurs among the groups. Each group responds with an activation $y_{j}$ (lines 16-17) and the loop started in line 19 repeats until a winner group is defined and updated. First, the group $J$ with higher activation is found (line 20). If this group was disabled (activation = -1), all groups were deactivated because of a reset sign, and a new group is created (lines 21-24). Otherwise, it is verified if the winner group is similar enough to the presented pattern (using the $\rho$ parameter). If not, the group is disabled and a reset occurs so that another group can be found (lines 25-33). If the winner group is considered similar enough to the input pattern, it is approximated to it (lines 36-40), the vectors updated are normalized (lines 42 and 43) and finally, the activations in layer $F_{1}$ are updated (lines 45-47).

The pattern recognition is done in a way very similar to the network training. The main difference is that there is no storage in the $F_{2}$ layer. Moreover, an adaptation of the $\rho$ parameter is done: it starts with an initial value next to 1 and is slightly reduced until a group is found in the $F_{2}$ layer.

The next section describes the simulations carried out with the proposed module.

\section{Simulations}
\label{sec:simulations}

The simulations aimed to reproduce the CSWL experiments available in the literature, following the methodology introduced by \cite{Yu2007} and further extended by others. Subsections from \secref{sec:learninguncertainty} to \secref{sec:multiplosreferentes} describe the CSWL experiments considered in this work and the respective simulations carried out with the proposed model. \secref{sec:dataset} describes the dataset used in the simulations. Notice that we employ the term ``Experiment'' to refer to the actual experiments carried out by \cite{Yu2007}, \cite{Yurovsky2013}, and \cite{Trueswell2013} with humans. The term ``Simulation'' refers to the simulations carried out with the proposed model, aiming to replicate each particular experiment. From \secref{sec:learninguncertainty} to \secref{sec:multiplosreferentes} we describe very briefly the considered CSWL experiments and their respective simulations. The details of the mentioned experiments are described in the \ref{sec:apCSWL}. Such subsections describing the simulations are divided into the following parts: first (i) a detailed description of the experiment conducted by the authors is presented, then (ii) the procedures used to simulate the experiments are described, and finally, (iii) the results produced by the simulations are presented in comparison with the results obtained in the original experiments. 

\begin{figure}[ht!]
\begin{minipage}{\linewidth}
    \includegraphics[width=0.97\linewidth]{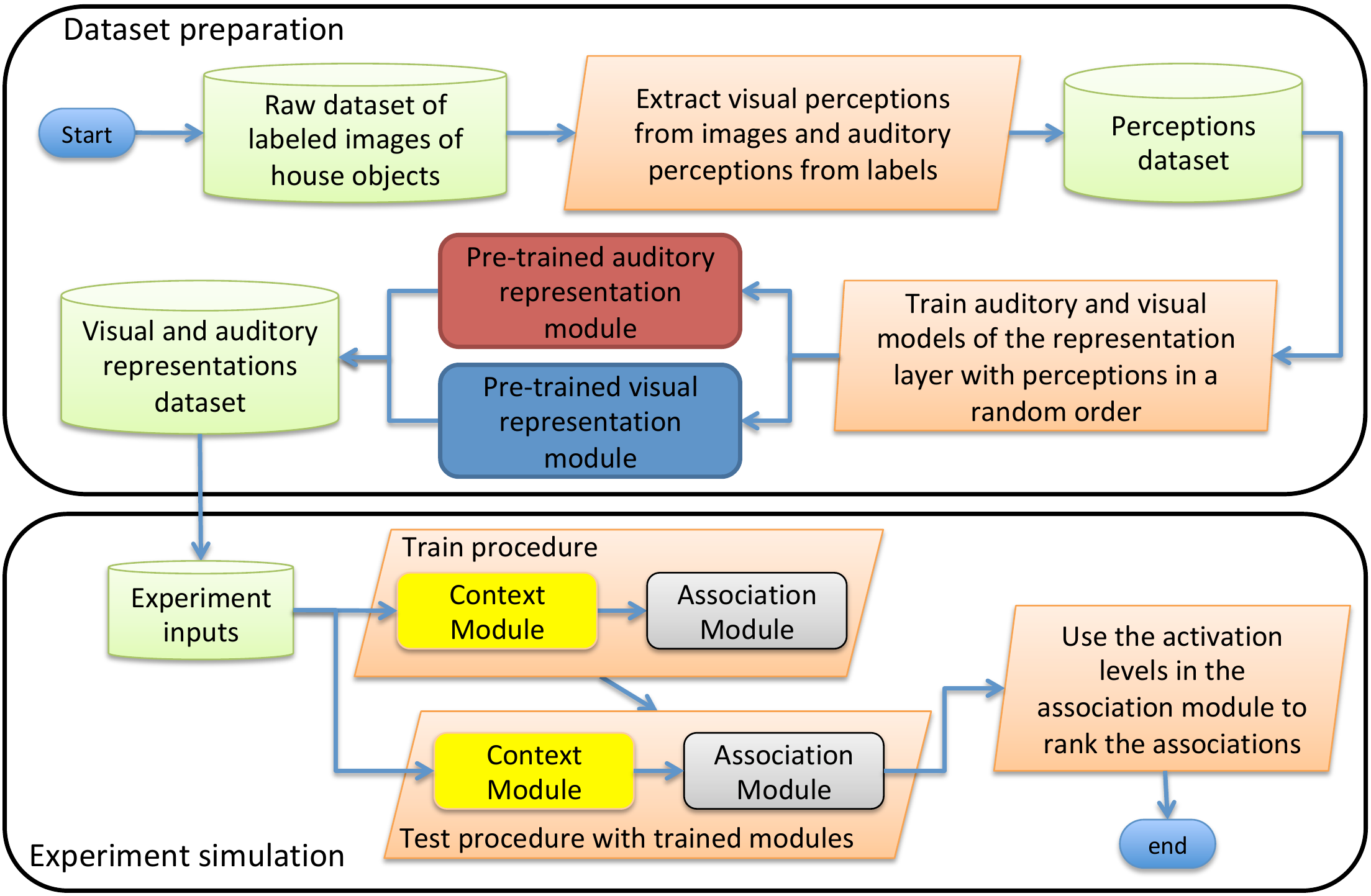}
    \captionof{figure}{Workflow of simulations: the process for generating the visual and auditory representations is illustrated in the \textbf{Dataset preparation} box. This process was executed only once and the same representations were used in all experiments. In the \textbf{Experiment simulation} box, is illustrated the process for simulating an experiment. This process was repeated for each experiment with different inputs, selected from the representations dataset according to the experiment design.}
    \label{fig:workflow}
\end{minipage}
\end{figure}

Furthermore, in \secref{sec:experimentocontexto}, the model with the adjusted set of parameters is evaluated in the last simulation, which aims to cover a part of the model that was not evaluated in the previous experiments: the Context Module and its role in providing the correct meaning for words with different meanings in different contexts. Since no work with this objective was found in the literature, an experimental design is firstly proposed to evaluate this ability in individuals, then, the results produced by the simulations of this experiment are presented. \figref{fig:workflow} illustrates the workflow for simulating the experiments. We start this subsections with the description of the parametric setup.

\subsection{Parameter Adjustment}
\label{sec:parameters}

    The parameters of each module of the proposed model were adjusted only once to minimize the differences between the results of all experiments and their respective simulations. The exploration of possible parameter values was made by employing a Latin Hypercube Sampling (LHS) \citep{Saltelli2009} and the best parameter set is presented in \tabref{tab:simulationParameters}.

The dataset used in the experiments and the way that each stimulus was presented to the model is detailed in the next section.

\begin{table}[ht!]
\begin{minipage}{\linewidth}
    \small
    \captionof{table}{Best parameter values obtained with the LHS adjustment.}
    \begin{tabular}{p{6.0cm}p{1.7cm}}
    \toprule
    \textbf{Parameter} & \textbf{Value}    \\
    \midrule
    \multicolumn{2}{p{7.5cm}}{\textbf{Visual Representation Module -- LARFDSSOM}}\\
    \midrule
    Activation threshold ($a_t$)         & 0.985    \\
    Lowest cluster percentage ($lp$)     & 0.15\%    \\
    Relevance rate ($\beta$)             & 0.10    \\
    Max competitions ($maxcomp$)         & 0.021$\times S$    \\
    Winner learning rate ($e_b$)         & $5 \times 10^{-4}$    \\
    Neighbors learning rate ($e_n$)      & $12\times 10^{-6} \times e_b$    \\
    Relevance smoothness ($s$)           & 0.007581760    \\
    Connection threshold ($c$)           & 0.50    \\
    \midrule
    \multicolumn{2}{p{7.5cm}}{\textbf{Auditory Representation Module -- LARFDSSOM}}\\
    \midrule
    Activation threshold ($a_t$)         & 0.935 \\
    Lowest cluster percentage ($lp$)     & 0.001\% \\
    Relevance rate ($\beta$)             & 0.10 \\
    Max competition ($maxcomp$)          & 2$\times S$\\
    Winner learning rate ($e_b$)         & 0.10 \\
    Neighbors learning rate ($e_n$)      & $14\times 10^{-6} \times e_b$ \\
    Relevance smoothness ($s$)           & 0.00394 \\
    Connection threshold ($c$)           & 0.50 \\
    \midrule
    \multicolumn{2}{p{7.5cm}}{\textbf{Context Module -- ART2 With Context}}\\
    \midrule
    Fixed weight in F1 ($a$)             & 10 \\
    Fixed weight in F1 ($b$)             & 10 \\
    Reset weight condition ($c$)         & 0.10 \\
    Winning Unit Activity in F2 ($d$)    & 0.9 \\
    A parameter to avoid division by zero ($e$)    
                                         & 0.0001 \\
    Noise suppression parameter ($\theta$)    
                                         & 0.0739221 \\
    Learning rate ($\alpha$)             & 0.8 \\
    Surveillance parameter    ($\rho$)   & 0.999 \\
    Number of Epochs       ($epochs$)    & 1 \\
    Number of Iterations  ($n_{iter}$)    & 1 \\
    Backpropagation context parameter   ($back$)    
                                    & 0.90 \\
    Context influences above reset mechanism    ($cw$)                                       & 0.0002 \\
    Winner Unit Activity in F2 for the context    ($d_{ctx}$)                         & 0.9 \\
    Context learning rate    ($\alpha_{ctx}$)    
                                    & 0.80 \\
    \midrule
    \multicolumn{2}{p{7.5cm}}{\textbf{Association Module -- LARFDSSOM}}\\
    \midrule
    Activity threshold ($a_t$)      & 0.999 \\
    Lowest cluster percentage ($lp$)& 17.5211\% \\
    Relevance rate ($\beta$)        & 0.870879 \\
    Maximum competition ($maxcomp$) & 10000 \\
    Winner learning rate ($e_b$)    & 0.465091 \\
    Neighbors learning rate ($e_n$) & 0.0134102$\times e_b$ \\
    Relevance smoothness ($s$)      & 1.31357 \\
    Connection threshold ($c$)      & 0.986745 \\
    \bottomrule
    \end{tabular}
    \label{tab:simulationParameters}
\end{minipage}
\end{table}

\subsection{The Real World Object Image and Label Dataset}
\label{sec:dataset}

In order to simulate the stimuli provided to the participants in the experiments of \cite{Yu2007}, we used 18 words of objects commonly found at home (\textit{armoire, bed, bowl, canister, chair, clock, computer, cooker, cup, desk, door, dresser, fork, knife, refrigerator, sofa, spoon}, and \textit{telephone}). In addition, 18 object images associated with these names were obtained from Google Image Search \textsuperscript{\textregistered}, using the respective word as the search term.

\figref{fig:crosssituacional} displays a sample of the object images collected. The complete dataset is available online (see footnote \ref{fn:github} on page~\pageref{fn:github}). This dataset was used in all simulations presented in the following subsections.

\subsection{Experiment 1: Word Learning Under Uncertainty}
\label{sec:learninguncertainty}

    \cite{Yu2007} evaluated the CSWL abilities of 38 undergraduate students dealing with slides containing pictures of unusual objects paired with pseudowords presented in auditory form. There were 3 groups of 18 pairs trained under different conditions concerning the number of labels and pictures that are presented (2 and 2, 3 and 3, or 4 and 4). Each subject was presented to 1word and 4 pictures and asked to choose the picture labeled by that word. The details of the experiments are in \ref{sec:apexp1}.

\subsubsection{Procedures for Simulation 1}

In the cross-situational experiments, the auditory stimuli, the sounds of the words formed a unique stream, thus, in each trial, a single auditory representation was created by chaining the representation of the sequence of phonemes of the words presented.

For example, assuming that the following four words are used in a test:  \textit{bed, chair, bowl} and \textit{fork}, the representation of the respective sequence of phonemes: \mbox{/b e d t \textesh{ } e \textschwa{ } b \textschwa{ } \textupsilon{ } f \textopeno{ } k/} formed the auditory input as described in \secref{sec:auditoryrepresentation}. On the other hand, in \cite{Yu2007} individuals could pay attention to each image at a time, observing them individually. Moreover, since there is not a strong correlation between the images, they make more sense when individually observed. Therefore, in our simulations, each image was represented individually, as described in \secref{sec:visualrepresentation}. Then, the input stimuli of a trial were constructed by paring the auditory stimulus with each one of the visual stimuli.

    For instance, in each trial of the 2x2 condition, two inputs were given for the model: one built by paring the auditory representation with the first image and another one built by pairing it with the second image.

    After the learning trials, analogously as in \cite{Yu2007} the testing consisted of presenting the sound of one word and four images. One of them is the correct association and the others are randomly chosen foils. The input stimuli are built similarly as in the training, with the only difference that now there is only one word, which its representation is paired with the representation of each one of the four images. To identify the association made by the model, each input pair is presented in a random sequence and the level of activity of the winner node in the association layer is registered. Then, the input pair that produced the highest activation is considered as the strongest association made by the model.

    The model was trained and tested 38 times, initialized with a different random seed, representing 38 different individuals. 

\subsubsection{Results of Experiment 1 and Simulation 1}

\begin{figure}[ht]
    \centering
    \includegraphics[width=0.9\linewidth]{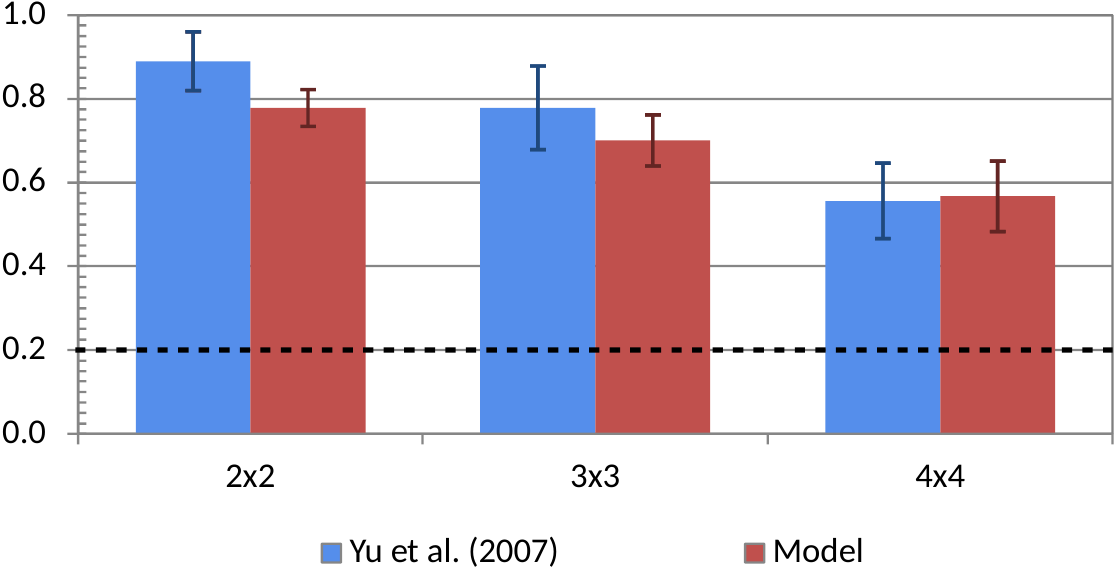}
    \caption{Experiments of \cite{Yu2007} in comparison with the results of our simulations. The strong horizontal dashed line indicates the probability of guessing by chance, while the error bars indicate the standard deviation.}
    \label{fig:resultsYu2007-Exp1}
\end{figure}

    \figref{fig:resultsYu2007-Exp1} shows that in the results obtained by \cite{Yu2007} in all conditions the individual have correctly guessed significantly more pairs ($0.889\pm0.07$, in condition 2x2, $0.778\pm0.10$, in 3x3 and $0.556\pm0.00$ in 4x4) then they would have by chance (1/4 = 0.25). Even in the most difficult condition (4x4), with 16 possible associations by trial, the individuals guessed on average 10 of the 18 pairs (0.55). The authors argue that humans are good at guessing the correct word-referent associations in situations of ambiguity and the results clearly show that the increase in the level of ambiguity inside the trials negatively affects the learning. This is confirmed by comparing the averages in conditions 2x2 and 4x4 in a $t$-test with a significance level of 1\%.

    Although there are visible differences, analogous conclusions can be drawn from the results of our simulations. The model could guess the correct associations better than chance and displayed a similar pattern of decay of learning as a function of the ambiguity inside trials ($0.778\pm0.044$, in condition 2x2, $0.700\pm0.061$, in 3x3, and $0.567\pm0.084$, in 4x4). The most significant difference is observed in condition 2x2, in which the model learns around 78\% of the pairs on average, while the individuals were able to learn about 89\%. Yet, the same $t$-test confirms that the learning rates in conditions 2x2 and 4x4 are statistically different.
    
\subsection{Experiment 2: Word Learning with More Than One Referent}
\label{sec:learningmorereferents}

    \cite{Yurovsky2013} experiments aimed to assess the behavior of individuals for words with two correct associations. A total of 48 students were tested in 18 word-referent pairs under 3 distinct conditions: each set of 6 words were associated with 1, 2, or none referents. In each of the 27 learning trials, the subject had to deal with 3 different word combinations. Then, each test consists of providing the subjects with 4 word-referent pairs to rank the most likely associations. The details of the experiments are in \ref{sec:apexp2}.

\subsubsection{Procedures for Simulation 2}

    In order to simulate the stimuli given to the participants of this experiment, the same 18 objects of the previous experiment were used. The six single words were: \textit{bed, chair, bowl, fork, door}, and \textit{canister} and presented together with their respective images. The six double words were: \textit{clock, computer, desk, refrigerator, sofa}, and \textit{cooker}, with their six respective images used as their first referents. The second referents of double words were images of different objects: respectively \textit{goblet, mat, mixer, crib, blender}, and \textit{shaker}. Finally, the six noise words were: \textit{spoon, telephone, knife, armoire, cup}, and \textit{dresser}.

    The paired input stimuli were built exactly as in the 4x4 condition of Experiment 1, and, in each testing trial, each one of the four testing words was selected (in a random order) and paired with each one of the four referents. The stimulus built for each pair was presented as input for the model and the activation of the winner node in the association layer was computed. Then, the activation levels were used to rank the pairs for computing the single, double, and either scores.

    This training and testing procedure was repeated 48 times with random initializations, representing the 48 participants.
    
\subsubsection{Results of Experiment 2 and Simulation 2}

    The results obtained by \cite{Yurovsky2013} (\figref{fig:resultsYurovsky2013-Exp2}) show that participants displayed a better-than-chance knowledge of the referents of single words ($0.454\pm0.264 > 0.25$), of one of the referents of double words ($0.698\pm0.210 > 0.5$), and even for both referents of double words ($0.301\pm0.146 > 0.17$), difference statistically verified by a $t$-test with a significance level of 1\%.

\begin{center}
\begin{minipage}{\linewidth}
    \includegraphics[width=0.9\linewidth]{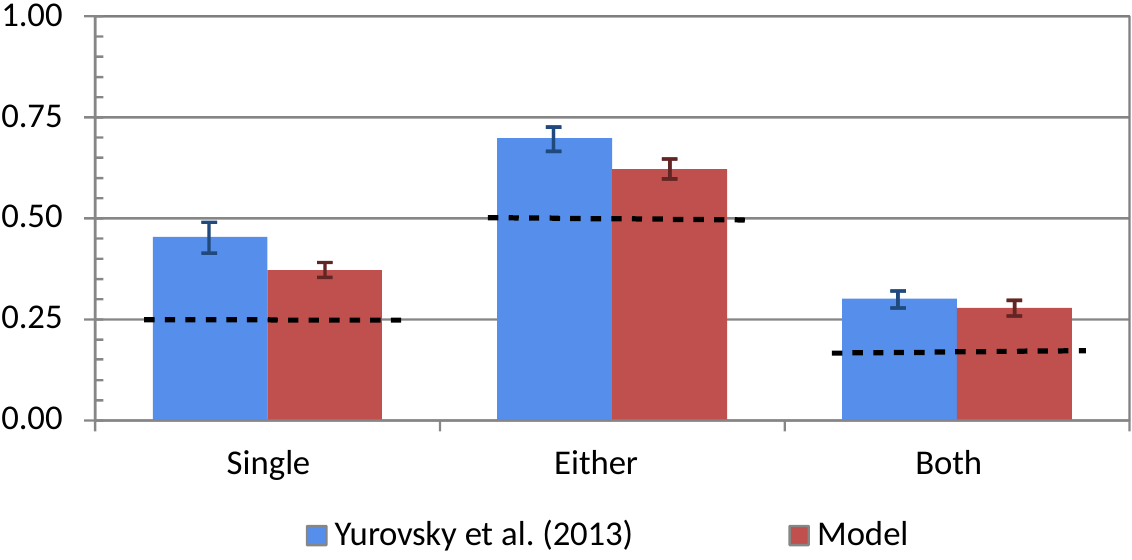}
    \captionof{figure}{Comparison of the results obtained by \cite{Yurovsky2013} with the result of the simulation with the proposed model in Experiment 2. Dashed lines indicate the chance levels of performance. The error bars indicate the Standard Error (SE), not Standard Deviation (SD), where $SE = SD/\sqrt{\text{\# of samples}}$.}
    \label{fig:resultsYurovsky2013-Exp2}
\end{minipage}
\end{center}

    \cite{Yurovsky2013} found that participants were significantly less likely to learn both referents of a double word than one referent of single words (t(47) = 3.68, p < .001). This suggests that two mappings composed of a single word and two different referents do not act like two independent mappings (two words and two different referents). This suggests the occurrence of some kind of competition for the mappings of a word.

    The same conclusions can be drawn from our simulations for single words  ($0.372\pm0.126>0.25$), one referent ($0.622\pm0.169>0.5$) and both referents ($0.278\pm0.135>0.17$) of double words. The model was also less likely to learn both referents of double words than one referent of single words (t(47) = 3.5267, p < .001).

    \cite{Yurovsky2013} also pointed out that, while this experiment allows concluding there is some kind of competition for the mappings, it is not clear which type of competition, local (within trials) or global (across trials), since both referents were shown in each trial. The next experiment addresses this issue.

\subsection{Experiment 3: Local vs Global Competition}
\label{sec:localglobalcompetition}

 \cite{Yurovsky2013} run experiments with 48 subjects who were trained with a single correct referent of double words. The individuals were asked to to the same test of the previous experimental. The details of the experiments are in \ref{sec:apexp3}.
 
\subsubsection{Procedures for Simulation 3}

Analogously as in Simulation 2, the six single words (\textit{bed, chair, bowl, fork, door, canister}) and double words (\textit{clock, computer, desk, refrigerator, sofa}, and \textit{cooker}) were the same, with their respective images. And the images of the same different objects (\textit{goblet, mat, mixer, crib, blender}, and \textit{shaker}) were used as the second meaning for double words. Noise words were not used and the testing procedure was kept the same of Simulation 2.

\subsubsection{Results of Experiment 3 and Simulation 3}

The results of this experiment (\figref{fig:resultsYurovsky2013-Exp3}), showed that, although participants knew all types of mappings above chance (single words: $0.400\pm0.247>0.25$; double words one referent: $0.580\pm0.277>0.5$; and both referents: $0.240\pm0.203>0.17$), they again showed better knowledge of single word referents than of both word referents (t(47) = 3.81, $p$ < 0.001).  This result suggests competition across trials.

\begin{center}
\begin{minipage}{\linewidth}
    \includegraphics[width=0.9\linewidth]{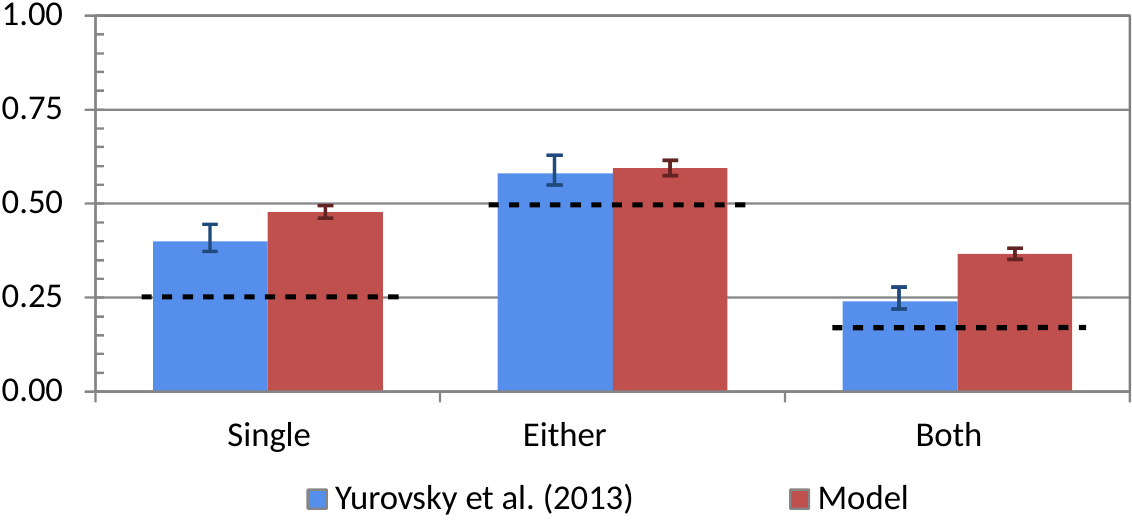}
    \captionof{figure}{Comparison of results obtained by \cite{Yurovsky2013} with the results obtained with the model in Experiment 3. Dashed lines indicate the chance levels of performance. The error bars indicate the Standard Error.}
    \label{fig:resultsYurovsky2013-Exp3}
\end{minipage}
\end{center}

    The simulations presented an analogous behavior with above-chance accuracy (single words: $0.478\pm0.115>0.25$; double words one referent: $0.594\pm0.142>0.5$; and both referents: $0.367\pm0.101>0.17$). The highest difference was observed for the recognition rate of both referents of double words, which could not be considered statistically equivalent to the results displayed by humans. In spite of that, the simulated participants also showed a better knowledge of single word referents than of both word referents (t(47) = 4.5613, $p$ < 0.0001), which also points to global competition.

\subsection{Experiment 4: Online vs Bach Learning}
\label{sec:globalcompetition}

\cite{Yurovsky2013} designed an experiment similar to Experiment 3 to assess the degree of globality of the competition process, i.e., they evaluated the influence of the temporal order of the individual trials upon accuracy. The details of the experiments are in \ref{sec:apexp4}.

\subsubsection{Procedures for Simulation 4}

    For the simulations, the same stimuli of the previous experiment were used for training and test. The only change was in the order of presentation of double words referents along the trials, which one of the referents of each double word was randomly chosen to be presented earlier, while the second referent was presented only after all presentations of the first referent.
    
\subsubsection{Results of Experiment 4 and Simulation 4}

    \figref{fig:resultsYurovsky2013-Exp4} shows the obtained results. Participants displayed similar results for single words ($0.450\pm0.300>0.25$) and for one referent of double words ($0.730\pm0.240>0.5$). However, they learned both referents of double words ($0.400\pm0.300>0.17$) as well as the referent of single words. Therefore, in contrast with previous experiments, the results did not show evidence of competition.

    A possible explanation given by \cite{Yurovsky2013}, is that while global competition protects old mappings from noisy information, local competition leverage prior mappings knowledge to speed up the acquisition of new mappings. 

\begin{center}
\begin{minipage}{\linewidth}
    \includegraphics[width=0.9\linewidth]{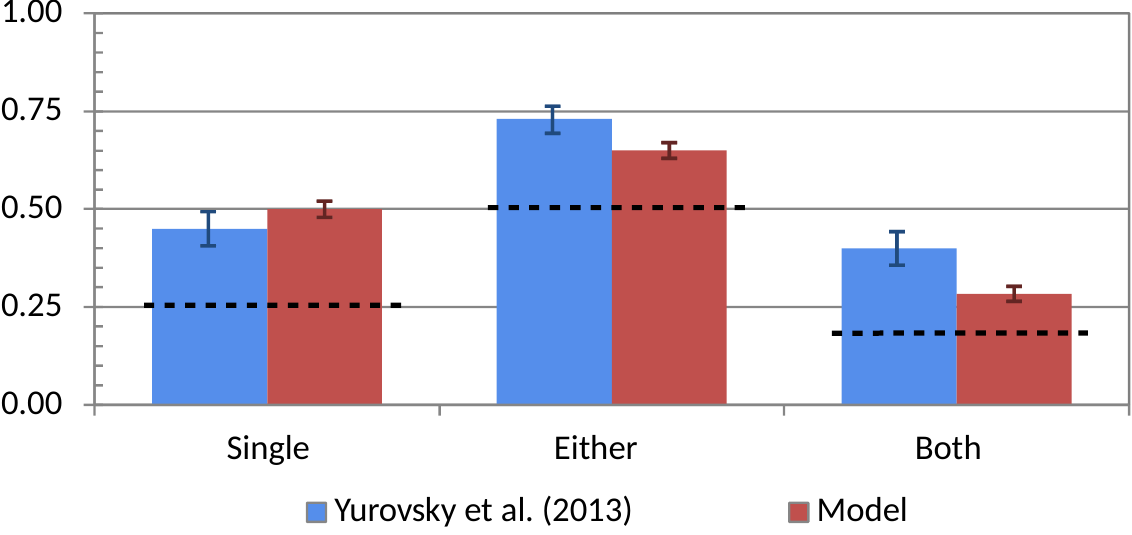}
    \captionof{figure}{Comparison of results obtained by \cite{Yurovsky2013} with the results obtained with the model in Experiment 4 for single, either and both words learning accuracy. Dashed lines indicate the chance levels of performance. The error bars indicate the Standard Error.}
    \label{fig:resultsYurovsky2013-Exp4}
\end{minipage}
\end{center}

	The simulations have shown similar results for single words ($0.500\pm0.146>0.25$) and for one referent of double words ($0.650\pm0.139>0.5$). However, differently from what participants have shown, the model did perform worse for both referents of double words ($0.283\pm0.129>0.17$). This was actually an expected result, since the model, in its present form, does not take any advantage of known mappings to speed up the acquisition of new mappings.

\begin{center}
\begin{minipage}{\linewidth}
    \includegraphics[width=0.9\linewidth]{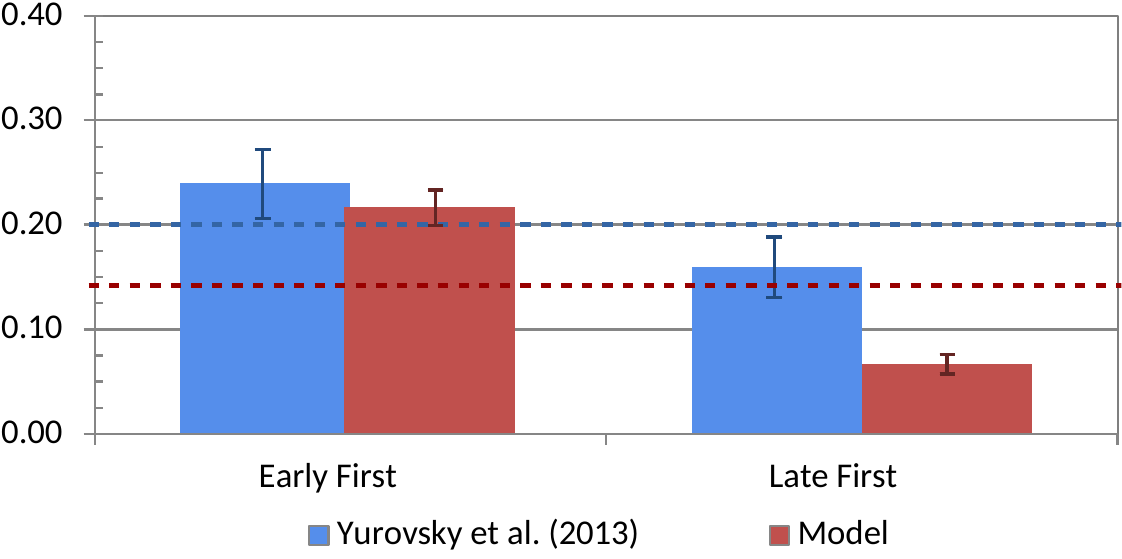}
   \captionof{figure}{Comparison of results obtained by \cite{Yurovsky2013} with the results obtained with the model in Experiment 4 for the frequency that Early and Late referents were ranked first. The error bars indicate the Standard Error and the dashed lines indicate the chance levels of performance in the experiment with humans (0.2) and in simulation (0.142).}
    \label{fig:resultsYurovsky2013-Exp4-b}
\end{minipage}
\end{center}

    Regarding the ordering factor, the results presented in \figref{fig:resultsYurovsky2013-Exp4-b} show that when participants picked up both correct referents for double words, they were slightly more likely ($t(47)=1.55, p=0.08$) to rank the early referent first ($0.24\pm0.23$) than the late referent ($0.16\pm0.20$). The model presented a similar pattern, though more strongly (early first: $0.217\pm0.12$; late first: $0.067\pm0.63$, $t(47)=7.8206, p<0.0001$).
    
    In the next section, we evaluate the capability of the model of reproducing the results of the experiments designed by \cite{Trueswell2013} to verify other learning aspects.
    
\subsection{Experiment 5: Statistical Association vs Propose-but-Verify}
\label{sec:multiplosreferentes}

    \cite{Trueswell2013} proposed the hypothesis ``Propose-but-Verify'' in which learning results from a one-trial procedure which links word-referent pairs that can be unlinked after opposite observations. To prove it, they designed experiments to verify if participants retain one or more association mappings for each word. They used 50 students to hear sentences an choose object referred by it. The individuals were supposed to learn association between phrases and images. The details of the experiments are in \secref{sec:apexp5}.

\subsubsection{Procedures for Simulation 5}

    For simulating the stimuli given to the participants in this experiment, the following 12 randomly chosen words were used among the 18 of Experiment 3: \textit{bed, chair, bowl, fork, door, canister, clock, computer, desk, refrigerator, sofa}, and \textit{cooker}. Also, the same 12 respective images were used as referents for these words.

    In each trial, the model was trained similarly as in previous experiments. The input stimuli were produced exactly as in previous experiments, combining the representation of the word with the representation of each referent, though now they were presented in the 1x5 condition, i.e., five combined input stimuli per trial.

    Differing from the previous simulations, this time, in order to match the procedure of the experiment, the level of activation of the winner node in the association layer was registered after each input stimulus. The referent that has resulted in the highest activation is considered as the choice of the model for the best association in the trial. The simulation was repeated 50 times with different random seeds to simulate the 50 participants.

\subsubsection{Results of Experiment 5 and Simulation 5}

    \figref{fig:resultsTrueswell2013-Exp5} shows the percentage of correct answers along the five learning cycles. As expected for a 1x5 condition, the average results suggest that the learning was more difficult than in previous experiments, though still viable. With the analysis of the growth of the learning curve, \cite{Trueswell2013} have shown that there was a significant increase in the accuracy throughout the learning cycles. The simulations have presented an analogous behavior. A $t$-test with 1\% of significance level confirms that both the participants and the model present an accuracy above chance at the last learning cycle.

\begin{center}
\begin{minipage}{\linewidth}
    \includegraphics[width=0.9\linewidth]{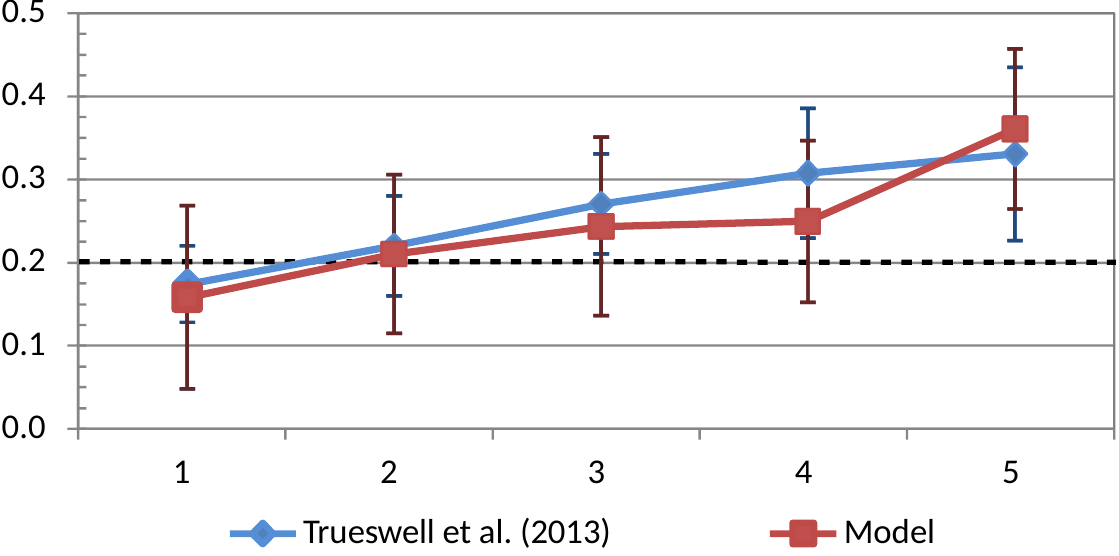}
    \captionof{figure}{Comparison of the accuracy growth through the learning cycles obtained by \cite{Trueswell2013} with the results obtained with the model in Experiment 5. Dashed lines indicate the chance level of performance. The error bars show a confidence interval of 95\%.}
    \label{fig:resultsTrueswell2013-Exp5}
\end{minipage}
\end{center}

    Since the previous result shows that learning still occurs in the 1x5 condition, the next step was to evaluate the hypothesis raised about the type of learning (Statistical Association vs Propose-but-Verify). As can be seen in \figref{fig:resultsTrueswell2013-Exp5-b}, the participants have identified the correct referent with an above-chance accuracy ($0.47\pm0.14$) only after assigning the correct referent in the previous cycle. When the participants missed the correct referent in the previous cycle, they seem to choose a random referent ($0.208\pm0.038 \simeq$ 0.20), presenting an accuracy near to 1 in 5 (randomly guessing). Therefore, even when a word has co-occurred before with the correct referent, participants show no sign of remembering it when they missed it in a previous cycle.

\begin{center}
\begin{minipage}{\linewidth}
    \includegraphics[width=0.9\linewidth]{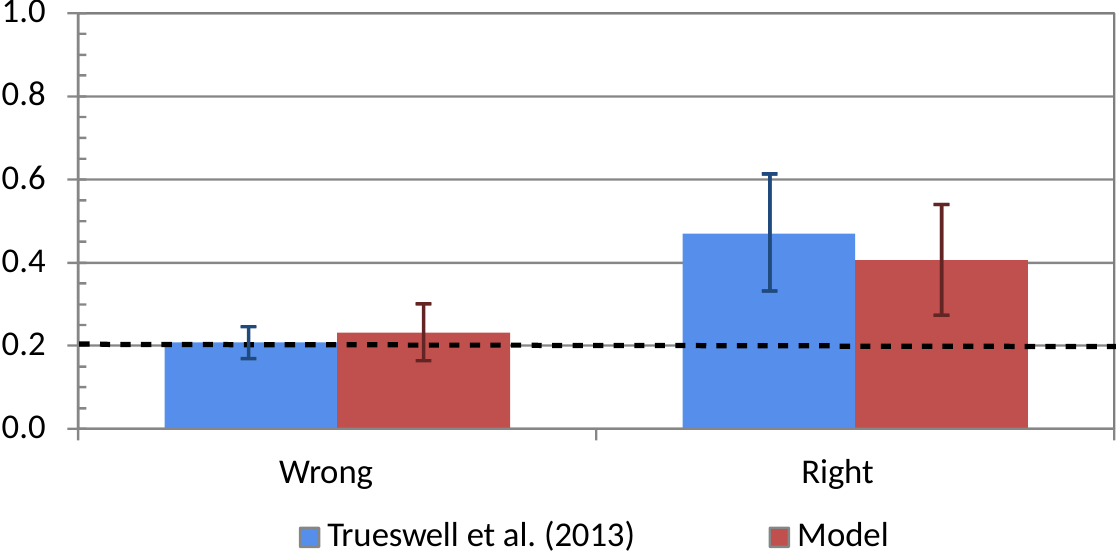}
    \captionof{figure}{Comparison of the results shown by \cite{Trueswell2013} with the results obtained with the model in Experiment 5. The ``Wrong'' label indicates the accuracy displayed when the wrong referent was chosen in the previous learning cycle and the ``Right'' label indicates the accuracy displayed when the correct referent was previously chosen. Dashed lines indicate the chance level of performance. The error bars show a confidence interval of 95\%}
    \label{fig:resultsTrueswell2013-Exp5-b}
\end{minipage}
\end{center}

    With this, \cite{Trueswell2013} conclude that participants did not retain multiple associations through the learning cycles. However, the proposed model has presented an analogous behavior, displaying an above chance accuracy only in the ``Right'' condition ($0.407\pm0.134$), while in the ``Wrong'' condition the results approach a random guess ($0.232\pm0.069$).

    We know, though, that the model can generate multiple hypotheses in each trial (up to five, in this case). This way, we are left with two possibilities: (a) the model did not generate multiple associations in each trial, or (b) the model did generate multiple associations, however, they were not strong enough to affect the accuracy significantly in the following cycle. In the model, the number of new associations generated in each trial is represented by the number of nodes created in the Association Module. Therefore, observing this value we can elucidate what has actually happened.

\begin{center}
\begin{minipage}{\linewidth}
    \includegraphics[width=0.9\linewidth]{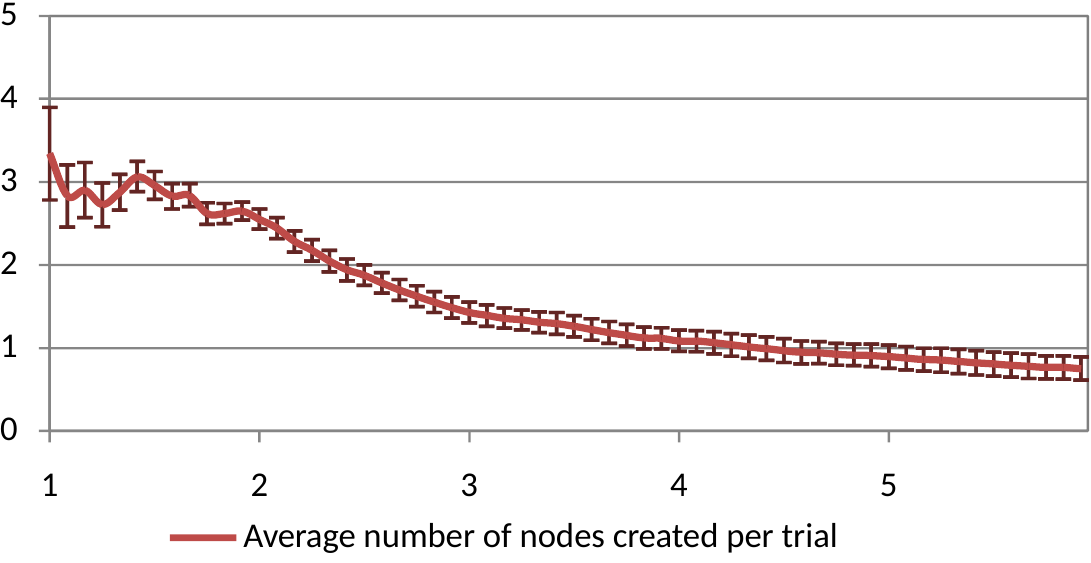}
    \captionof{figure}{Number of nodes created in the Association Module after each trial (1-5), through the learning cycles in Experiment 5. The error bars show the standard deviation.}
    \label{fig:resultsBassani2014-Exp5-c}
\end{minipage}
\end{center}

    \figref{fig:resultsBassani2014-Exp5-c} shows the evolution of the number of nodes created in the Association Module in each trial, through the learning cycles. In the first cycle, the model creates  about 3 nodes per trial on average, and this number decays through the learning cycles, reaching a value below 1 in the last cycle. This is an expected behavior since, after some learning, the model has already created the associations required to represent most mappings. This indicates that hypothesis (b) is the correct one. The model generates multiple associations, though not as many as possible, however, they are not strong enough to affect the accuracy in the next learning cycle.

    We argue that two factors may explain the results observed by \cite{Trueswell2013} without disregarding the hypothesis of multiple associations. One is that global competition may insert noise in the associations formed, degrading weak associations (seen only once). The other factor is that in the experimental design of \cite{Trueswell2013}, the number of incorrect associations is computed from the second to the fifth cycle, when the number of associations created by trial may have decreased, as our simulations suggest. Therefore, in our model, a chance level accuracy for words that were incorrectly associated in the previous cycle is not a result of retaining a single association hypothesis.

\subsection{Experiment 6 Design: The Role of The Context in Word Disambiguation}
\label{sec:experimentocontexto}

	In all previous simulations, the context module was active and functional. The results obtained in those simulations show that it does not interfere with the learning in the evaluated conditions. However, the role of the context module itself was not directly evaluated. When a word has different meanings, they are usually employed in different situations (contexts). Therefore, our hypothesis is that learning the context in which words are used can help the model to learn their different meanings. 
    
    In order to evaluate this, we designed the following experiment, based on the 1x5 condition proposed by \cite{Trueswell2013}: The stimuli are composed of two lists of six words (A and B), sharing exactly one word, the ambiguous label (AL), that is associated with a different referent in each list, i.e., in list A the label AL is associated with the referent RA while in list B it is associated with a different referent, RB. 

    The training should be carried out in six cycles of 14 trials each: the odd cycles are done with words from list A (including AL), while the even cycles are done with words from list B (including AL). Therefore, the cycles are intercalated in the form: (A, B, A, B, A, B). This training aims to induce the creation of two different contexts associated with the words in each list. Since the context changes slowly, the associations created with words in the same list tend to be similar, since they are consecutively presented.

    The testing procedure aims to verify if the recovered referent for the AL matches the context induced by the stimuli previously given as input, and how many stimuli are necessary to induce the context. Thus, in order to induce the context, six trials in the condition 1x4 are done using words from one of the lists (excluding AL), before testing the association of AL, also in condition 1x4, with both referents (RA and RB) and two other randomly chosen referents (lures), one from list A and the other from list B. The context inducting conditions are: 3a+3b, 3b+3a, 4a+2b, 4b+2a, 5a+1b and 5b+1a. The condition 3a+3b, for instance, indicates that three trials with labels and referents from list A were presented, followed by three trials with labels and referents from list B. In this condition, it is expected that the context of list B (late list) is induced, thus, the correct association to be retrieved is with referent RB. 

    In each testing trial, three results are possible: (i) the referent of the list presented later is chosen (expected association); (ii) the referent from the list presented earlier is chosen; or (iii) one of the lures is chosen. The prior for the situation (i) is 0.25 (one in four referents), while the prior for selecting one of both associations, (i) or (ii), is 0.5 (two in four).

\subsubsection{Procedures for Simulation 6}

    The words chosen to simulate the Experiment 6 were: \textit{armoire, snake, dog, cat, cheese, trap} and \textbf{\textit{mouse}} for the first list, and \textit{speaker, printer, computer, notebook, monitor, keyboard}, and \textbf{\textit{mouse}} for the second list. Note that the ambiguous label, AL, is the word \textit{mouse}. The referents for both lists consisted again of images download from Google Image Search \textsuperscript{\textregistered}, using the respective word as the search term. The two referents for the AL consisted of an image of the animal (RA), and an image of the computer device (RB).

    Training and testing were done according to the experimental design described above, and the input stimuli combining the word representation with the representation of each referent were produced exactly as in previous experiments. The simulation was repeated 48 times with different random seeds to simulate 48 participants. 

\subsubsection{Results of Simulation 6}

    The obtained results are shown in \figref{fig:ResultBassani2014-Exp6}. In conditions 3+3 (3a+3b and 3b+3a) and 4+2 (4a+2b and 4b+2a) the context was effective to induce the recovery of the correct referent with a high accuracy (respectively, $0.937\pm0.167$ and $0.739\pm0.252$), with an expected decay in accuracy from conditions 3+3 to 4+2. In conditions 5+1 (5a+1b and 5b+1a), however, the accuracy falls to $0.5\pm0.145$, which means that the contextual information is not enough to induce the correct association. The model seems to have difficulty choosing between the two possible referents RA and RB, though it can easily discard the lures. A $t$-test with 1\% of significance level confirms that these results are different between them and that they are above chance.

\begin{center}
\begin{minipage}{\linewidth}
    \includegraphics[width=0.9\linewidth]{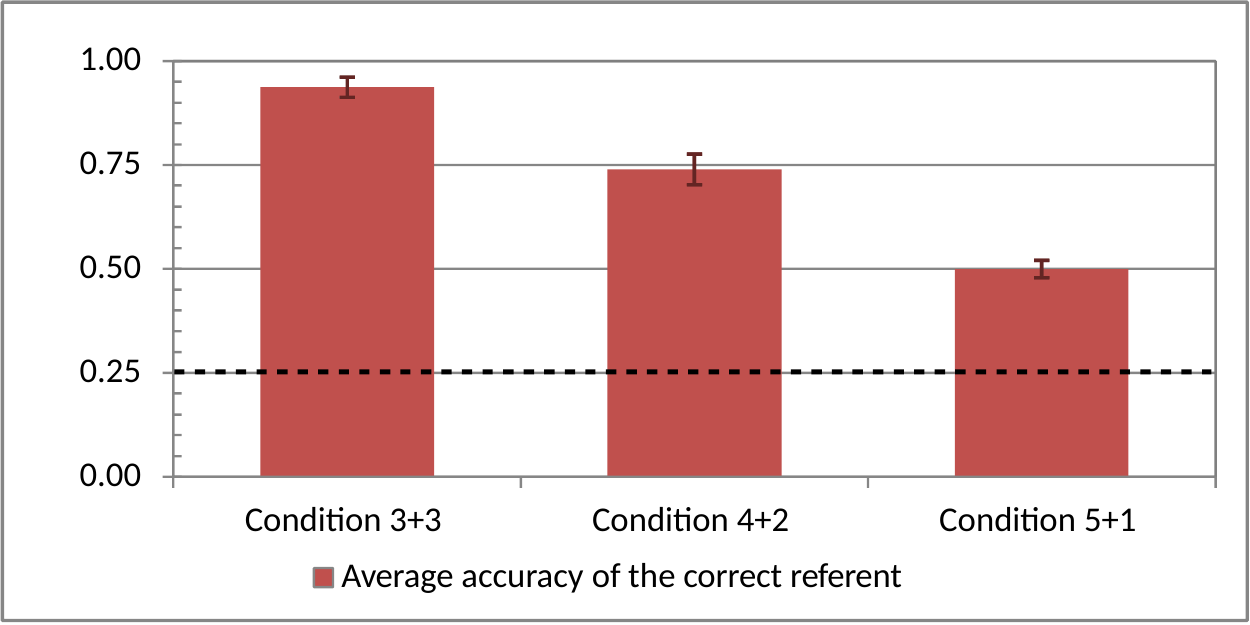}
    \captionof{figure}{Accuracy of the model in choosing the referent induced by the context in each condition: In 3+3, the last three words induce the desired context; in 4+2, the last two words induce the desired context; and in 5+1, only the last word induces the desired context.}
    \label{fig:ResultBassani2014-Exp6}
\end{minipage}
\end{center}

    These results emphasize the role of the context, showing that it can help to recover the correct meaning for ambiguous words.

\section{Discussion}
\label{sec:conclusaoexperimentos}

	The experimental paradigm of the cross-situational word learning has shown to be a very useful tool for evaluating the hypothesis about the mechanisms that allow us to learn word-referent associations. The model described in this article has been proposed considering pieces of evidence accumulated in the studies of psycholinguistics and neurolinguistics, organized in a modular architecture that allows us to better understand and communicate about the functions required for word-referent associations.
    
    The results obtained in Experiment 1 are similar in terms of accuracy to the results of the models evaluated by \cite{Yu2012}. However, this article also considered other conditions not evaluated in previous works and introduces advances in terms of model architecture in comparison with previous models. One improvement is the use of a Time-Varying Self-Organizing Map (LARFDSSOM) as the point of connection between the visual and auditory layers. In previous models, this was done via associative connections trained by Hebbian learning. In the proposed model, LARFDSSOM learns the correlations between the different input dimensions from the co-variations observed in the input data by the means of its relevance learning mechanism. This is similar to Hebbian learning, however, it has other useful features such as the topological representation of input data and the activation levels produced by the nodes, which allowed us to model the CSWL experiments. Moreover, the same kind of map is used in different levels of the architecture, which seems to be more plausible.

    The proposed architecture is far from being a complete model of cross-situational word learning. It is, however, a step towards a model that allows us to simulate and evaluate hypotheses about the mechanisms behind this characteristic of human nature. The fact that it can deal with real-world inputs, images for referents and text or sound for words, gives it an enormous flexibility for simulating more accurately several types of experiments carried out with human beings. We notice that in several conditions its association accuracy is a little below the accuracy of humans. For instance, in the 2x2 condition of Exp.1, humans can reach 89\% of accuracy while the model achieved only 78\%. This could be due to the fact that handcrafted features extractors were used to represent images and words in the first layer. This can be further improved by employing modern representation techniques such as word embeddings \citep{Mikolov2013} and convolutional neural networks, already shown to work well in combination with LARFDSSOM \citep{MEDEIROS2019} and to achieve human level performance in certain image classification scenarios \citep{He2015}. 

    In spite of that, the simulations show that the model is suitable for replicating most of the experiments considered in this work, allowing us to draw similar conclusions. However, in Experiment 4, it seems that it is much easier for humans to learn a second referent of a double word after having learned the first referent than learning both simultaneously, while this was not the case for the proposed model. We evaluate that this happens because the model cannot take advantage of other known associations within a trial to reduce ambiguity. For instance, in a 4x4 condition, if a human participant knows three of the four associations, he might easily guess the fourth association by disregarding words and referents in the other three. The model, otherwise, only strengthens the current stronger association for each pair. Therefore, in future work, the model should be modified to take this into account.

    In spite of that, the main conclusions obtained by \cite{Yu2007,Yurovsky2013}, and \cite{Trueswell2013} with their experiments, including those in Experiment 4, could also be drawn from the simulation results, summarized below:

\begin{description}
    \item[Exp.1:] The model was able to simulate the remarkable ability of participants in learning associations between labels and referents in different levels of ambiguity, including the fact that it decays with the increase of ambiguity;

    \item[Exp.2:] The model was able to replicate the greater difficulty that participants present to learn two referents of the same label than to learn only one referent;

    \item[Exp.3:] Global competition is also the most relevant type of interference that degrades the learning of the model as seems to be the case with the individuals;

    \item[Exp.4:] The model employs online learning instead of batch learning, which matches the type of learning identified by \cite{Yurovsky2013} in their experiments. However, the model does not benefit from the knowledge of previously known mappings when forming new associations, which prevented it from reproducing part of the results observed by the authors.

    \item[Exp.5:] Though more difficult, learning still occurs in the high ambiguity condition 1x5, and the model could replicate this fact accurately in each one of the five learning cycles.
\end{description}

	It is also worth noting that in Experiment 5, the simulations allowed us to verify how many associations were created in each trial, which has shown that it is possible for a method that makes multiple associations to achieve the results observed by \cite{Trueswell2013}, in opposition to the authors' assumption. This is an example of how such kind of modeling can be useful in the evaluation of new hypotheses.

    Another improvement of the current architecture, in relation to previous models, was the introduction of the context module. With this, we could evaluate how context can be used to retrieve the correct meaning of ambiguous words. Therefore, the results obtained in the simulation of Experiment 6 can and should be evaluated in experiments with human beings to verify how well the model predicts the effects of the context in the disambiguation of words meanings.
    
    Finally, the proposed model may be applied in the proposition and test of new hypotheses and experimental paradigms, contributing to understanding the mechanisms involved in word learning, and can be used as a component for developing agents that learn natural language. Still, it is important to emphasize that although this model was developed taking into consideration the current knowledge provided by neuroscience and cognitive psychology, it is a high level computational model and may not reflect the real learning and representation mechanisms that occur in the brain.

\section*{Acknowledgment}

    The authors would like to thank CNPq (Conselho Nacional de Desenvolvimento Científico e Tecnológico) and FACEPE (Fundação de Amparo à Ciência e Tecnologia do Estado de Pernambuco) for supporting project \#APQ-0880-1.03/14.

\section*{References}
\bibliography{references}

\appendix
\section{More Information on CSWL Experiments}
\label{sec:apCSWL}

\subsection{Exp.1: Word Learning Under Uncertainty}
\label{sec:apexp1}

\cite{Yu2007} evaluated the CSWL abilities of 38 undergraduate students. The stimuli provided consisted of slides containing 2, 3 or 4 pictures of unusual objects paired respectively with 2, 3 or 4 pseudowords presented in auditory form. These artificial words were generated by a computer program using standard phonemes in the English language, the native language of the participants. In this case, there were 54 label-referent pairs formed by single and unique objects randomly chosen and divided into three groups of 18 pairs, which were used in three different training conditions.

The distinct training conditions differ only in the number of labels and referents simultaneously presented to the test subjects. In the 2x2 condition, two labels and two pictures were presented in each trial; in the 3x3 condition, three labels and three pictures were presented in each trial; and, in the 4x4 conditions, four labels and four pictures were presented in each trial. During the trials, there was no indication of which label goes with each picture. However, in the underlying label-referent mappings it is guaranteed that an individual label was present in a training trial, if and only if the referent was also present. \figref{fig:crosssituacional} illustrates a 4x4 condition.

In the test procedures, the participants were told that multiple words and pictures would co-occur in each trial and that their task was to figure out across trials which word went with which picture. They were not told that there was one referent per word. After training in each condition, subjects received a four-alternative forced-choice test of learning, in which, they were presented with 1 word and 4 pictures and asked to indicate the picture named by that word. The target picture and the 3 foils were all drawn from a set of 18 training pictures.

\subsection{Exp.2: Word Learning with More Than One Referent}
\label{sec:apexp2}

\cite{Yurovsky2013} performed a series of experiments to evaluate the behavior of individuals when there are two correct associations. In the first experiment, 48 grad students were evaluated, also with 18 word-referent pairs. However, the pairs were split in different conditions: six words were associated with a single referent (\textit{single words}), six words were associated with two referents (\textit{double words}), and the last six words had no associated referents (\textit{noise words}). 

The \textit{single words} play the same role as those in the previous experiment, always co-occurring with their referents in each trial. The \textit{double words}, however, co-occur with both referents in each trial. Since both single and double words co-occur six times with their referents, the total number of occurrences is the same for both types of words. The noise words occur with the same frequency for all referents, thus, they are not consistently mapped to any referent. They serve only for producing an equal number of words in all the trials.

Each trial consists of presenting the stimuli in the 4x4 condition. From a total of 27 trials (\figref{fig:exemplo-Yurovsky2013-Exp2}), in two of them the stimuli were composed of four single words; in 14 trials the stimuli were composed of two single words, one double word, and one noise word; and in 11 trials the stimuli were composed of two double words, and two noise words. This way, although in all trials there were always four words and four referents, the mapping structure varied considerably across the trials, and in only two of them it consisted exclusively of one-to-one mappings.

\begin{center}
\begin{minipage}{\linewidth}
    \includegraphics[width=0.90\linewidth]{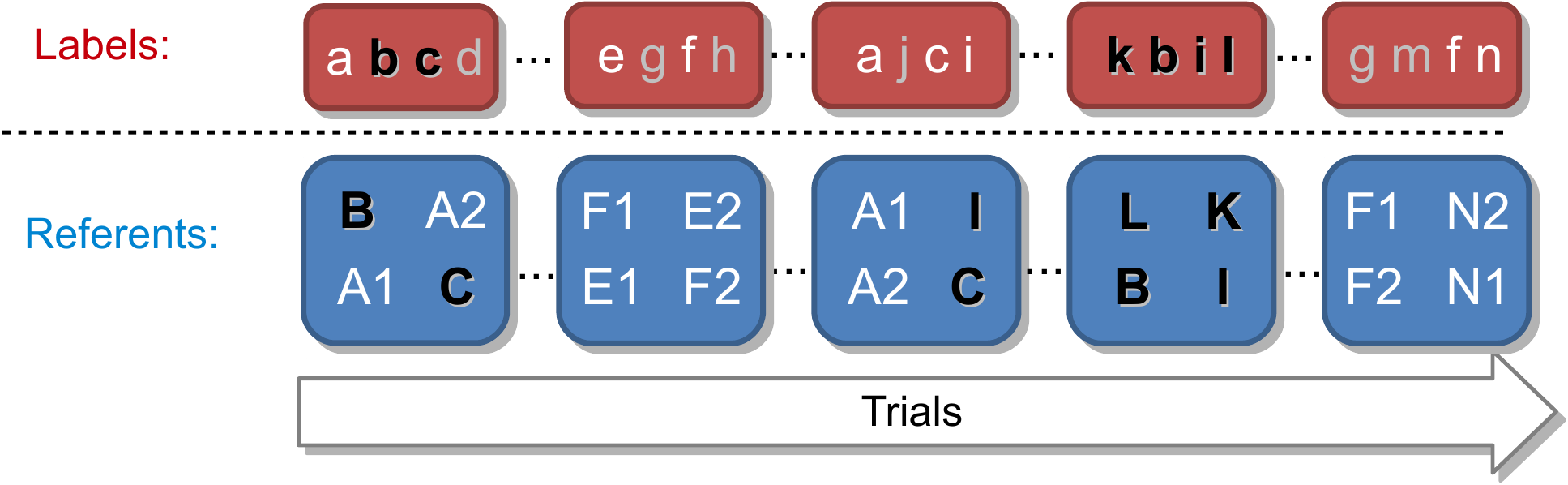}
    \captionof{figure}{Structure of Experiment 2. The lowercase letters represent words and the uppercase letters represent referents. Single words are in bold (ex.: b-B and c-C), double words are in white (ex.: a-A1 and a-A2, f-F1 and f-F2), and noise words are in gray (ex.: d and g).}
    \label{fig:exemplo-Yurovsky2013-Exp2}
\end{minipage}
\end{center}

After the learning trials, the learning rates of each individual were evaluated similarly as in \cite{Yu2007}. Every single word was presented with its referent and three other randomly chosen referents and each double word was presented with both their referents and two other randomly chosen referents.

The individuals were asked to rank the four objects from the most to the least likely meaning of the word. To compute the scores of single words, one correct guess is computed when the correct referent was ranked first. For double words, two types of scores were computed: a \textit{Double} score is computed when the participant ranks both correct referents (in either order) in the first and second positions, and a \textit{Either} score is computed when the participant ranks one of the correct referents in the first position and an incorrect referent in the second position.

\subsection{Exp.3: Local vs Global Competition}
\label{sec:apexp3}

    To evaluate if global or local competition has occurred, \cite{Yurovsky2013}, in this experiment, another 48 participants were exposed to only one correct referent of double words in each trial, while the testing procedure was the same of the previous experiment. If only local competition during training was occurring, then the participants of this trial should be able to learn both referents of double words as well as they learn the referent of single words. Otherwise, global competition was occurring. The stimuli were presented as illustrated in \figref{fig:exemplo-Yurovsky2013-Exp3}. Noise words were not necessary for this experiment since two single words and two double words were presented in each trial with their respective referents (4x4 condition).

\begin{center}
\begin{minipage}{\linewidth}
    \includegraphics[width=0.90\linewidth]{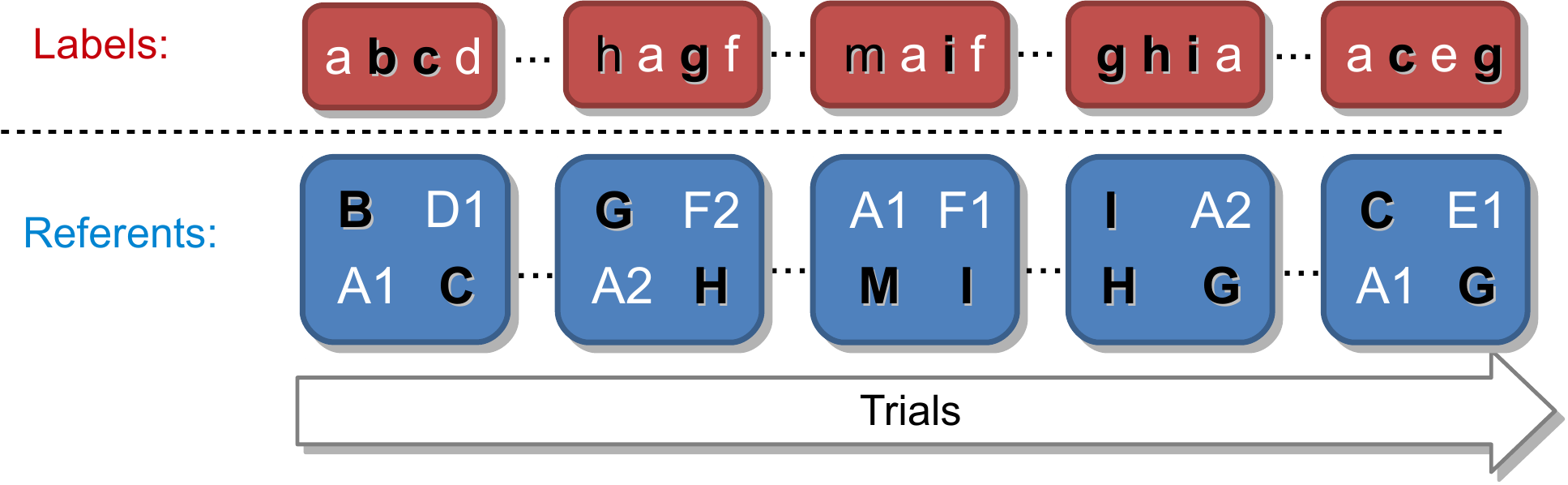}
    \captionof{figure}{Structure of Experiment 3. Differently from Experiment 2, in this experiment, only one correct referent of double words is presented in each trial. The co-occurrence frequency of correct associations was the same of Experiment 2.}
    \label{fig:exemplo-Yurovsky2013-Exp3}
\end{minipage}
\end{center}

\subsection{Exp.4: Online vs Bach Learning}
\label{sec:apexp4}

\cite{Yurovsky2013} conjectured that if the competition is primarily global, and occurs only after all training information has been accumulated (batch learning), there should be no effect of the temporal order of the individual trials. However, if global competition emerges trial-by-trial (online learning), and does not interact with other local mappings within a trial, then it is expected that a decrement of the accuracy will be observed for the second referents of double words presented later in relation to the knowledge of referents presented earlier. 

\cite{Yurovsky2013} designed an experiment to evaluate this, with the organization shown in \figref{fig:exemplo-Yurovsky2013-Exp4} for a new group of participants. Notice that this experiment is similar to Experiment 3. However, one of the referents of each double word is randomly chosen to be presented earlier, while the second referent is presented only after all co-occurrences with the first referent have been carried out. Notice also that both referents have the exact same frequency of co-occurrence with their respective double word.

\begin{center}
\begin{minipage}{\linewidth}
    \includegraphics[width=0.90\linewidth]{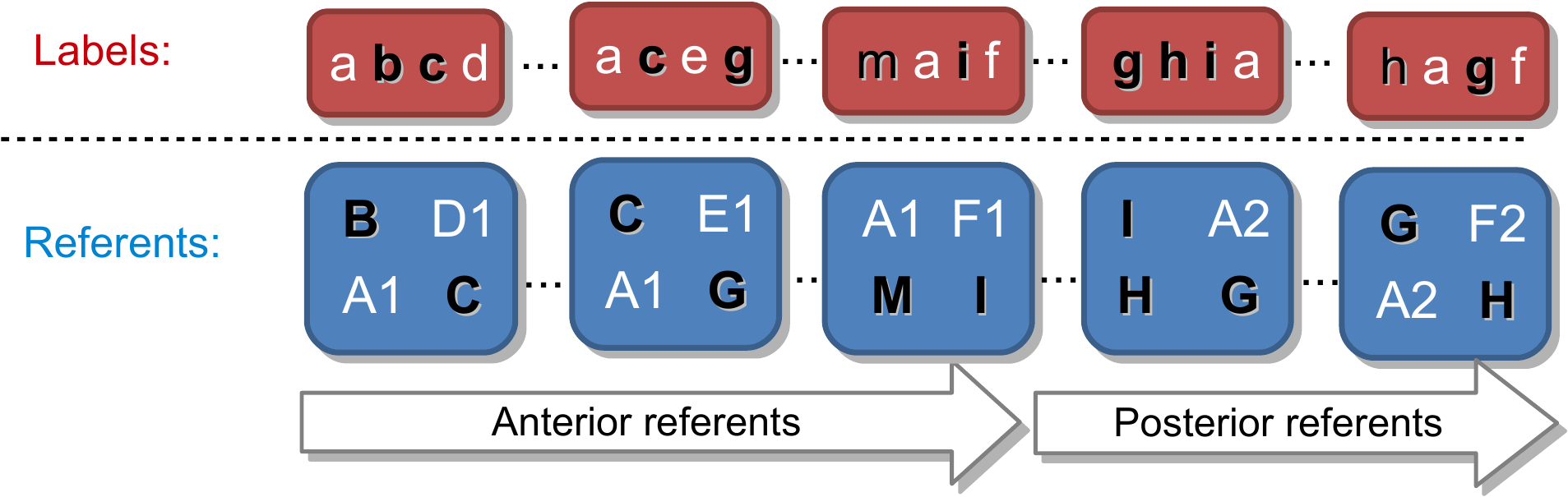}
   \captionof{figure}{Structure of Experiment 4. Differently from Experiment 3, in this experiment, one of the referents of double words is presented first (A1), while the other is presented in later trials (A2). The co-occurrence frequency of correct associations was the same of Experiments 2 and 3.}
    \label{fig:exemplo-Yurovsky2013-Exp4}
\end{minipage}
\end{center}

\subsection{{Exp.5}: Statistical Association vs Propose-but-Verify}
\label{sec:apexp5}

Although the results of previous experiments suggest that learning under such conditions derives from some kind of statistical-associative learning mechanism, as the one the proposed model employs, \cite{Trueswell2013} suggest the hypothesis that learning is instead the product of a one-trial procedure in which a single hypothesized word-referent pairing is made in one shot and retained across learning instances, being abandoned only if a subsequent observation fails to confirm the pairing. The authors called this hypothesis ``Propose-but-Verify''.

In order to test this, \cite{Trueswell2013} designed experiments to explicitly verify if participants retain a set of association mappings for each word or if they keep a single conjecture about the association.

In each of the trials designed by \cite{Trueswell2013}, five images were used as referents, while the auditory stimuli consisted of phrases such as ``Oh! look, a ...!'' with one label (condition 1x5). In total, 12 artificial words were used as labels and 12 images of objects were used as referents. In such a scenario, there is a high degree of uncertainty about the correct referent. 

The trials were divided into five learning cycles. In each cycle, each word was presented once in a random order. The other four cycles are repetitions of the first cycle in the same order. 

Fifty undergrad students participated in the tests. They were instructed that, after hearing the phrase, they should click on the object referred by the phrase. 

Since the participants were tested in every trial, this allowed the authors to register the evolution of the learning rates of the individuals after each learning cycle. The rationale is that if participants store only one association, and the referent is not the correct one, then when finding the same word in a subsequent trial, they should choose randomly between the available referents and should not show any bias for the correct referent, since there should be no trace in memory of such association. A bias for the correct referent should be observed if the participants can keep track of multiple possible associations.

\section{Numeric Representation of Phonemes}
\label{sec:apnumericrepphon}

\begin{table*}[hb!]
	\renewcommand{\tabcolsep}{0cm}
	\footnotesize
	\centering
	\caption{Correspondence between IPA and ARPAbet symbols and the respective numeric representation of each phoneme.}
	\begin{tabular}{lc >{\centering\arraybackslash}m{1.5cm} | >{\centering\arraybackslash}m{0.9cm}		>{\centering\arraybackslash}m{0.9cm}		>{\centering\arraybackslash}m{0.9cm}		>{\centering\arraybackslash}m{0.9cm}		>{\centering\arraybackslash}m{0.9cm}		>{\centering\arraybackslash}m{0.9cm}		>{\centering\arraybackslash}m{0.9cm}		>{\centering\arraybackslash}m{0.9cm}		>{\centering\arraybackslash}m{0.9cm}		>{\centering\arraybackslash}m{0.9cm}		>{\centering\arraybackslash}m{0.9cm}		>{\centering\arraybackslash}m{0.9cm}}
	\toprule
	Phoneme & IPA & ARPAbet & \multicolumn{12}{c}{Numeric Representation} \\
	\midrule																				
	f\underline{a}ther	&	\textscripta	&	AA	&	1	&	0.5	&	1	&	-1	&	0	&	0	&	0	&	0	&	0	&	0	&	0	&	0	\\
	\underline{a}t	&	\ae	&	AE	&	1	&	-0.5	&	-1	&	-1	&	0	&	0	&	0	&	0	&	0	&	0	&	0	&	0	\\
	b\underline{u}t. sof\underline{a}	&	\textturnv. \textschwa	&	AH	&	0.67	&	0	&	-1	&	-1	&	0	&	0	&	0	&	0	&	0	&	0	&	0	&	0	\\
	\underline{o}ff	&	\textopeno	&	AO	&	0.33	&	1	&	1	&	1	&	0	&	0	&	0	&	0	&	0	&	0	&	0	&	0	\\
	h\underline{ow}	&	a\textupsilon	&	AW	&	0	&	0.5	&	0	&	0	&	0	&	0	&	0	&	0	&	0	&	0	&	0	&	0	\\
	m\underline{y}	&	a\textsci	&	AY	&	0	&	0	&	-0.5	&	0	&	0	&	0	&	0	&	0	&	0	&	0	&	0	&	0	\\
	r\underline{e}d	&	\textepsilon 	&	EH	&	0.33	&	-0.5	&	-1	&	-1	&	0	&	0	&	0	&	0	&	0	&	0	&	0	&	0	\\
	h\underline{er}. cow\underline{ar}d	&	\textrhookrevepsilon. \textrhookschwa	&	ER	&	0.33	&	0	&	1	&	0	&	0	&	0	&	0	&	0	&	0	&	0	&	0	&	0	\\
	b\underline{i}g	&	\textsci	&	IH	&	-0.67	&	-0.5	&	-1	&	-1	&	0	&	0	&	0	&	0	&	0	&	0	&	0	&	0	\\
	b\underline{ee}	&	i	&	IY	&	-1	&	-1	&	1	&	-1	&	0	&	0	&	0	&	0	&	0	&	0	&	0	&	0	\\
	b\underline{oy}	&	\textopeno\textsci	&	OY	&	0	&	0	&	0	&	0	&	0	&	0	&	0	&	0	&	0	&	0	&	0	&	0	\\
	sh\underline{ow}	&	o\textupsilon	&	OW	&	-0.33	&	1	&	1	&	1	&	0	&	0	&	0	&	0	&	0	&	0	&	0	&	0	\\
	s\underline{ay}	&	e\textsci	&	EY	&	-0.33	&	-1	&	1	&	0	&	0	&	0	&	0	&	0	&	0	&	0	&	0	&	0	\\
	sh\underline{oul}d	&	\textupsilon	&	UH	&	-0.67	&	0.5	&	-1	&	0	&	0	&	0	&	0	&	0	&	0	&	0	&	0	&	0	\\
	y\underline{ou}	&	u	&	UW	&	-1	&	1	&	1	&	1	&	0	&	0	&	0	&	0	&	0	&	0	&	0	&	0	\\
	\midrule																													
	\underline{b}uy	&	b	&	B	&	0	&	0	&	0	&	0	&	1	&	-1	&	1	&	-1	&	1	&	-1	&	-1	&	-1	\\
	\underline{ch}air	&	t\textesh	&	CH	&	0	&	0	&	0	&	0	&	0.27	&	-1	&	-1	&	0	&	-1	&	-1	&	-1	&	-1	\\
	\underline{d}ay	&	d	&	D	&	0	&	0	&	0	&	0	&	0.45	&	-1	&	1	&	-1	&	1	&	-1	&	-1	&	-1	\\
	\underline{th}at	&	\dh	&	DH	&	0	&	0	&	0	&	0	&	0.64	&	-1	&	-1	&	1	&	1	&	-1	&	-1	&	-1	\\
	\underline{f}or	&	f	&	F	&	0	&	0	&	0	&	0	&	0.82	&	-1	&	-1	&	1	&	-1	&	-1	&	-1	&	-1	\\
	\underline{g}o	&	g	&	G	&	0	&	0	&	0	&	0	&	-0.27	&	-1	&	1	&	-1	&	1	&	-1	&	-1	&	-1	\\
	\underline{h}ouse	&	h	&	HH	&	0	&	0	&	0	&	0	&	-1	&	-1	&	-1	&	1	&	0	&	-1	&	-1	&	-1	\\
	\underline{j}ust	&	d\textyogh	&	JH	&	0	&	0	&	0	&	0	&	0.45	&	-1	&	-1	&	0	&	1	&	-1	&	-1	&	-1	\\
	\underline{k}ey	&	k	&	K	&	0	&	0	&	0	&	0	&	-0.27	&	-1	&	1	&	-1	&	-1	&	-1	&	-1	&	-1	\\
	\underline{l}ate	&	l	&	L	&	0	&	0	&	0	&	0	&	0.45	&	-1	&	-1	&	-1	&	1	&	-1	&	-1	&	1	\\
	\underline{m}an	&	m	&	M	&	0	&	0	&	0	&	0	&	1	&	1	&	-1	&	-1	&	1	&	-1	&	-1	&	-1	\\
	k\underline{n}ee	&	n	&	N	&	0	&	0	&	0	&	0	&	0.45	&	1	&	-1	&	-1	&	1	&	-1	&	-1	&	-1	\\
	si\underline{ng}	&	\textrtailn	&	NG	&	0	&	0	&	0	&	0	&	-0.27	&	1	&	-1	&	-1	&	1	&	-1	&	-1	&	-1	\\
	\underline{p}ay	&	p	&	P	&	0	&	0	&	0	&	0	&	1	&	-1	&	1	&	-1	&	-1	&	-1	&	-1	&	-1	\\
	\underline{r}un	&	r. \textturnr	&	R	&	0	&	0	&	0	&	0	&	0.27	&	-1	&	-1	&	-1	&	1	&	-1	&	1	&	-1	\\
	\underline{s}ay	&	s	&	S	&	0	&	0	&	0	&	0	&	0.45	&	-1	&	-1	&	1	&	-1	&	-1	&	-1	&	-1	\\
	\underline{sh}ow	&	\textesh	&	SH	&	0	&	0	&	0	&	0	&	0.27	&	-1	&	-1	&	1	&	-1	&	-1	&	-1	&	-1	\\
	\underline{t}ake	&	t	&	T	&	0	&	0	&	0	&	0	&	0.45	&	-1	&	1	&	-1	&	-1	&	-1	&	-1	&	-1	\\
	\underline{th}anks	&	\texttheta	&	TH	&	0	&	0	&	0	&	0	&	0.64	&	-1	&	-1	&	1	&	-1	&	-1	&	-1	&	-1	\\
	\underline{v}ery	&	v	&	V	&	0	&	0	&	0	&	0	&	0.82	&	-1	&	-1	&	1	&	1	&	-1	&	-1	&	-1	\\
	\underline{w}ay	&	w	&	W	&	0	&	0	&	0	&	0	&	1	&	-1	&	-1	&	-1	&	1	&	1	&	-1	&	-1	\\
	\underline{y}es	&	j	&	Y	&	0	&	0	&	0	&	0	&	-0.09	&	-1	&	-1	&	-1	&	1	&	1	&	-1	&	-1	\\
	\underline{z}oo	&	z	&	Z	&	0	&	0	&	0	&	0	&	0.45	&	-1	&	-1	&	1	&	1	&	-1	&	-1	&	-1	\\
	mea\underline{s}ure	&	\textyogh	&	ZH	&	0	&	0	&	0	&	0	&	0.27	&	-1	&	-1	&	1	&	1	&	-1	&	-1	&	-1	\\
	-- & silent	&	\#	&	0	&	0	&	0	&	0	&	0	&	0	&	0	&	0	&	0	&	0	&	0	&	0	\\
    \bottomrule
	\end{tabular}
	\label{tab:PhoneticSymbolsRepresentation}
\end{table*} 

The numeric representation of the auditory data was constructed following a procedure similar to the one described in \cite{Araujo2010}. First, each word is converted to its respective phonetic representation according to the CMU Pronouncing Dictionary \citep{Lenzo2007}, in which, each phoneme is represented by its ARPAbet symbol (\tabref{tab:PhoneticSymbolsRepresentation}). Then, each phoneme is translated into a vector of 12 real values ranging from -1 to +1, according to its place of pronunciation in the International Phonetic Alphabet (IPA) charts for vowels (4 features) and consonants (8 features). For example, the word ``ball'' is converted as follow:

 \[\text{ball} \rightarrow \text{B AO L}\rightarrow \left[
\begin{smallmatrix}
0	&	0	&	0	&	0	&	1	&	-1	&	1	&	-1	&	1	&	-1	&	-1	&	-1	\\
1	&	0.5	&	1	&	-1	&	0	&	0	&	0	&	0	&	0	&	0	&	0	&	0	\\
0	&	0	&	0	&	0	&0.45	&	-1	&	-1	&	-1	&	1	&	-1	&	-1	&	1
\end{smallmatrix}\right]\]
\end{document}